\documentclass[
]{ceurart}

\usepackage{listings,adjustbox,placeins,xurl}
\begin{document}

\copyrightyear{2026}
\copyrightclause{Copyright for this paper by its authors.
  Use permitted under Creative Commons License Attribution 4.0
  International (CC BY 4.0).}

\conference{}

\title{Generalizable Multiscale Segmentation of Heterogeneous Map Collections}

\author[1,2]{Remi Petitpierre}[%
orcid=0000-0001-9138-6727,
email=remi.petitpierre@epfl.ch,
]
\address[1]{EPFL, École Polytechnique Fédérale de Lausanne. Digital Humanities Laboratory (DHLAB). CM 1 148, Station 10, 1015 Lausanne, Switzerland.}
\address[2]{EPFL. Urban Sociology Laboratory (LASUR).}



\begin{abstract}
Historical map collections are highly diverse in style, scale, and geographic focus, often consisting of many single-sheet documents. Yet most work in map recognition focuses on specialist models tailored to homogeneous map series. In contrast, this article aims to develop generalizable semantic segmentation models and a shared class ontology.  We introduce Semap, a new open benchmark dataset comprising 1,439 manually annotated patches designed to reflect the variety of historical map documents. We also evaluate a segmentation workflow that combines procedural data synthesis with multiscale integration. Under the harmonized benchmarks considered here, this workflow attains the strongest performance among the evaluated approaches and viably learns from heterogeneous map data. The results indicate that segmentation performance remains largely stable across map collections, scales, geographic regions, and publication contexts. By providing benchmark data, a shared ontology, and an evaluation of generic segmentation across heterogeneous maps, this work contributes toward integrating the long tail of cartographic archives into historical geographic studies.
\end{abstract}


\maketitle

\section{Introduction}\label{sec:1}

Historical maps contain dense information on historical geography, representing a valuable source for economic history, climate science, and urban studies. Applications include measuring urban sprawl, guiding forest conservation policies, and modeling the historical development of road networks \cite{BGEQN6UN,LZ5GNULF,BMBEHXRD}. Over the last decades, heritage institutions worldwide have undertaken the digitization of their cartographic collections, resulting in hundreds of thousands of map images made available online \cite{IKTPDJBV}. Consequently, specific approaches have been developed to segment and recognize the information displayed on map images. 

Historical maps tend to be stylistically diverse, each tending to adopt a distinct graphical form. Additionally, relatively few annotated data are available, which constitutes a challenge for supervised computer vision. Consequently, the majority of research projects focus on large, visually uniform datasets. Map series, like topographic maps or city atlases, often cover sizable territories in a graphically congruent way. Therefore, they are considered a sound target for automated extraction workflows, as a reasonable amount of dedicated training data is usually sufficient to train an average-sized specialist neural network.

Concentrating on specific datasets naturally undermines the transfer potential for models. Moreover, the annotated data are rarely reused on other projects, and the results themselves may not apply to distinct cartographic contexts. The low reuse potential also undermines training data release, which perpetuates the data scarcity confronting the field.

In this article, we provide two distinct responses to these challenges. First, we assemble and publish Semap, a diverse dataset for semantic segmentation tasks, which constitutes one of the primary contributions of this work. This dataset can serve as a benchmark for comparing approaches across map scales, production contexts, and geographic coverage, and as a source of pretraining data. Second, we evaluate a generic map-recognition workflow built from established segmentation components, procedural data synthesis, and multiscale inference.

The experiments support the viability of a diversity-driven approach to map recognition: a model trained with diverse data can retain strong performance across distinct cartographic contexts. This opens useful perspectives for the long tail of cartographic archives. Whereas map series are convenient because they cover large territories consistently, they represent only a fraction of historical map collections. The long tail comprises the much larger and more diverse set of maps outside such series. Incorporating these sources is important for realizing the potential of cartographic data at scale, and ensuring the provenance diversity, temporal depth, and geographic coverage on which computational map studies are based.

The remainder of the article is structured as follows. First, we review the literature related to the segmentation of historical maps. Second, we present Semap, a generic dataset for map segmentation, and introduce the procedural data-synthesis method. The third section describes the semantic-segmentation workflow, including multiscale inference. The fourth section compares its performance with earlier approaches under harmonized benchmarks. Finally, the last section displays and discusses several qualitative segmentation results.

\FloatBarrier

\section{Related work}\label{sec:2}

The last decade has witnessed rapid progress in historical map recognition, driven by convolutional and Transformer-based neural networks. In a seminal work, Uhl et al. \cite{P32JKYH2} introduced a patch-based classifier. However, the first application of semantic segmentation to maps was presented by Oliveira et al. \cite{63LS594U}, who demonstrated the efficacy of UNet \cite{ESMCSY4S} architectures and ResNet \cite{UYYWZK49} encoders to segment historical cadastres. Petitpierre \cite{CGK4FENK} extended the approach to diverse medium-scale city maps. In parallel, Heitzler and Hurni \cite{U23QD4TD,LVXMMF5T} evidenced the scalability of CNN-based methods, as well as their relevance to process even smaller-scale maps, by extracting thousands of buildings and city blocks from the Siegfried Swiss topographic map series.

At the same time, Chiang et al. \cite{ZVQFDMWJ} addressed the extraction of railroads from the USGS map series. Because the reliable and uninterrupted detection of linear features poses a significant challenge, roads quickly emerged as a primary extraction target \cite{QDTIFG72,JHEVKQ8S,JZMFGI49}. Jiao et al. \cite{CBST88UR} presented a comprehensive survey of road extraction methods. While Hosseini et al. \cite{R6RP2WMJ} recommended reverting to coarser, yet more reliable, patch-based classifiers, Jiao et al. introduced pioneering approaches that rely on synthesized training samples for segmenting roads in the Siegfried maps \cite{LZ5GNULF,TI8VL2E4}. To address the issue of line discontinuity, Xia et al. \cite{BSB33Q7P} proposed an end-to-end vector framework that implements Transformer-based line segment detection \cite{S7WQIMQ4}.

In urban-historical contexts, segmentation has proven to be useful for extracting building footprints and identifying urban extents \cite{PVXGKXCR}. Chazalon et al. \cite{MN8LTGQS} and Chen et al. \cite{FP55GPVF} concentrated on closed shape extraction from the Paris atlases. By contrast, Petitpierre et al. \cite{4H4XGM2Y} focused on the extraction of distinct land classes from Swiss Napoleonic cadastral maps, using HRNet, a high-resolution CNN-based model for edge filtering \cite{UZ9LB7IU}, and OCRNet, a Transformer-based model for contextual segmentation \cite{IBI9YNCQ}. Other studies similarly attempted to extract building footprints from Franciscean cadastres with CNNs \cite{XCDIEM9C,TV7XH2BW}. These technologies have also attracted interest in environmental history, for the study of land use change \cite{PU5SZPTU,7XVY2CL5}, wetland monitoring \cite{VPHRFJP5}, and surface mine mapping \cite{AWGGZTC7}.

Several recent studies on map recognition still rely on CNN-based frameworks \cite{ETSAEFRC}. In a benchmark paper, Chen et al. \cite{KPIDQQ4F} argued that Vision Transformers \cite{7U3E4BRJ} are unable to produce reliable segmentation masks. Similarly, Wu et al. \cite{PKWUZHSU} reported the inferior performance of Segformer \cite{VYLMUW8M} relative to UNet. However, other context-aware Transformer-based architectures, like OCRNet \cite{QWHQBJDN}, have achieved performance gains compared to convolutional UNets. Xia et al. \cite{ZYYEWSAX} retained a UNet backbone, but demonstrated the potential of Swin encoders \cite{AGXDA7XV} to address the complex task of railway segmentation. Wu et al. \cite{PKWUZHSU} proposed a multi-head cross-attention module that leverages spatio-temporal cues to segment aligned maps, treating the extraction of map series as a joint segmentation task. In a similar attempt to exploit spatial context, Arzoumanidis et al. \cite{R943LAEU,D33PL94I} introduced self-constructing graph convolutional networks (SCGCN), reporting encouraging results. Other promising approaches include contrastive learning \cite{ZYYEWSAX} and the adaptation of large vision models such as Segment Anything \cite{LECIPW5W,8Q9EIGGG}. 

The performance of semantic segmentation models is largely contingent on the quality of the training data \cite{DMHFPAJU}. To alleviate this time-consuming task, some scholars proposed to propagate semantic annotations across time and throughout map series \cite{5BC2XM8P,P4FDQUVK}. By contrast, several studies have highlighted the potential of synthetic data pretraining to improve the model’s robustness to diverse geographic configurations and graphic noise. This strategy has yielded promising results in road segmentation \cite{LZ5GNULF,9BVZMKVZ,MUL2WJ9M}, and map text detection tasks \cite{S2XGZJE4,RZI93QMS,Y84567UY}. Other works have shown that adversarial generative models can enrich the diversity of map samples through style transfer \cite{6IKIYPW2,VU7QJ78P}. Nevertheless, they also highlighted the difficulty of generating credible synthetic historical data directly from GIS-derived information or segmentation masks \cite{W9UTP692}.

\FloatBarrier

\section{Data}\label{sec:3}

\subsection{Semap Dataset, a new annotated dataset for generic map recognition}\label{sec:3.1}

Here, class ontology refers to the shared semantic organization used to translate heterogeneous map content into consistent segmentation annotations. These pixel-level annotations assign semantic classes to the depicted map features and regions, as well as to non-geographic map components. Semap defines six mutually exclusive classes—background, boundary, built, non-built, water, and road network—which are described below.

Therefore, at present, one of the principal obstacles to the development of semantic segmentation techniques for historical maps is the scarcity of generic ground-truth annotations with which to pretrain models. One open-source dataset that tends in this direction is the Historical City Maps Semantic Segmentation Dataset (HCMSSD, \cite{4YXKU2KM}. This dataset, which exhibits a variety of content and style, contains 330 annotated map patches corresponding to city maps of Paris (HCMSSD Paris), and 305 additional patches covering 182 distinct cities in 90 countries, collected from 32 heritage institutions and digital libraries (HCMSSD World). Each patch measures 1,000\,×\,1,000 pixels. The annotations encompass five semantic classes; the first four pertain to geographic content: built (e.g., buildings, walls, porticoes), road network (e.g., streets, squares, walkways, bridges), water (e.g., rivers, seas, lakes, canals), and non-built—which covers agricultural and natural land as well as green spaces (e.g., parks, gardens). The fifth class, background, captures map components that are not directly geographic, such as document borders, legends, ornaments (e.g., cartouches), and peripheral layout components (e.g., compasses, graphical scales).

However, this article addresses an even wider variety of maps, both in terms of content and scale, in order to consider all types of archived maps—from insurance plans to world maps, including chorographic maps. This diversity is made possible by the Aggregated Database on the History of Cartography (ADHOC), a dataset containing a balanced sample of 99,715 digitized maps from 16 heritage collections in Europe and the United States \cite{PJC6ZP3W}.

The Semantic Segmentation Map Dataset we present here, abbreviated Semap, consists of 1,439 manually annotated map samples. Semap is a composite dataset balanced to match ADHOC images while capitalizing on existing annotated data. Aside from a curated subset of 356 patches from the HCMSSD, it also includes 78 samples extracted from Napoleonic or renovated cadastres \cite{2ED5JES3,HXWMWXM9,3HCXCCDR} and three from Paris city atlases \cite{MN8LTGQS}. These data are augmented with an additional 1,002 newly annotated samples measuring 768\,×\,768 pixels\footnote{The choice of this lower resolution is determined by the architecture of the segmentation model, and memory constraints. Besides, it enables a slightly shorter annotation time per sample, thereby allowing more samples to be annotated and a greater variety of maps to be included in the training set.}, drawn from ADHOC images. The overall composition of Semap is detailed in Table~\ref{tab:a1}, in Appendix.

\begin{figure}[pos=htbp,align=\centering]
\includegraphics[width=\linewidth]{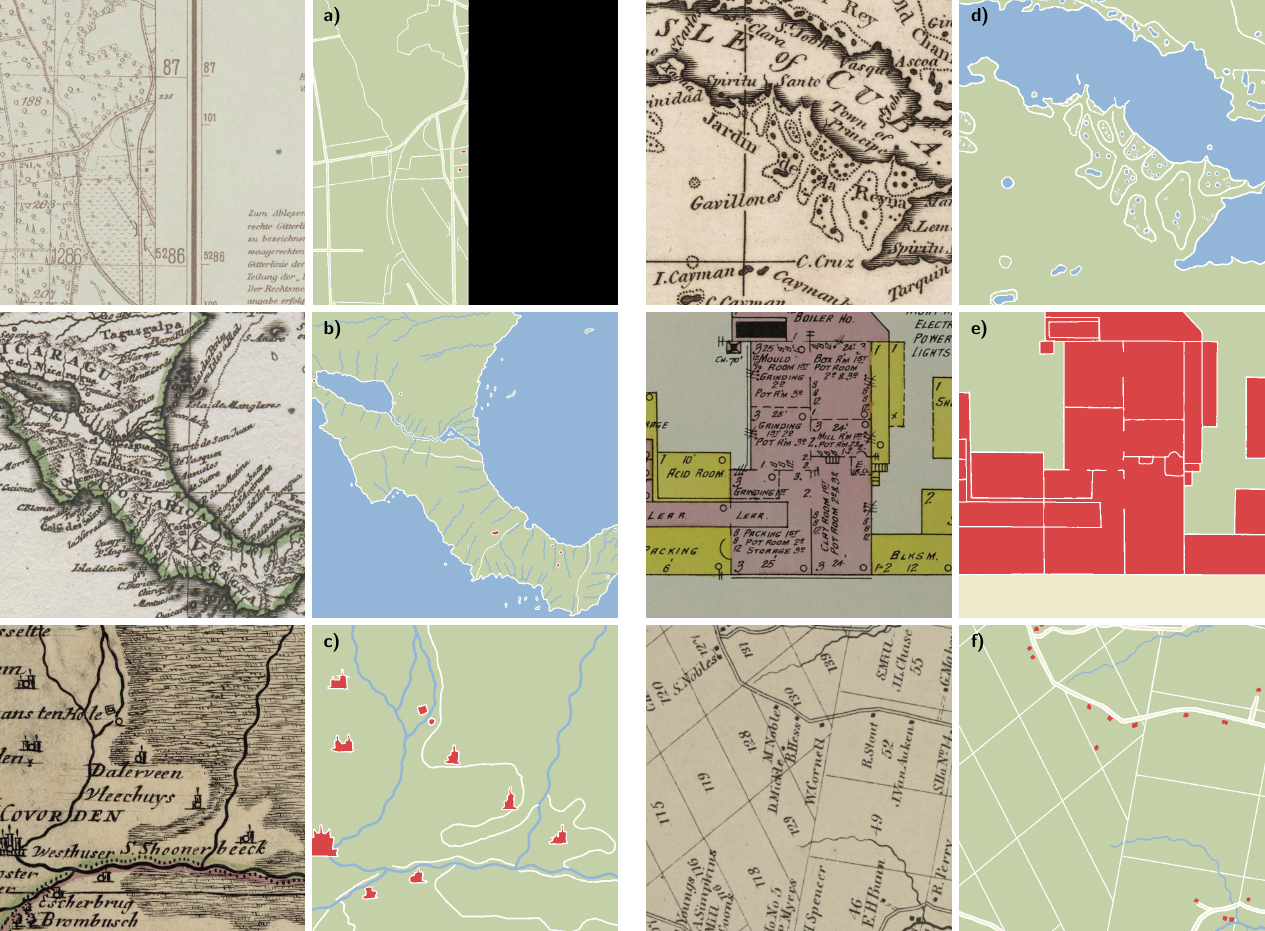}
\caption{Manually annotated training samples. Real map crops (left) with corresponding annotations (right). The annotation color code follows the legend of Fig.~\ref{fig:2}.}
\label{fig:1}
\end{figure}

The addition of a sixth semantic class, boundary, enriches the annotations by encoding edges between adjacent objects or land-cover regions. In this paper, boundary is treated as an auxiliary semantic class within an area-segmentation task. Topology-preserving instance extraction and generic vectorization are not evaluated. Introducing this sixth class required updating all samples from HCMSSD and related datasets to include the supplementary class. In total, the annotation process demanded approximately 400 hours of manual work. Sample annotations from the resulting Semap dataset, illustrating the diversity of the training data, are shown in Figure~\ref{fig:1}.

\subsection{Procedural data synthesis}\label{sec:3.2}

\begin{figure}[pos=htbp,align=\centering]
\includegraphics[width=.95\linewidth]{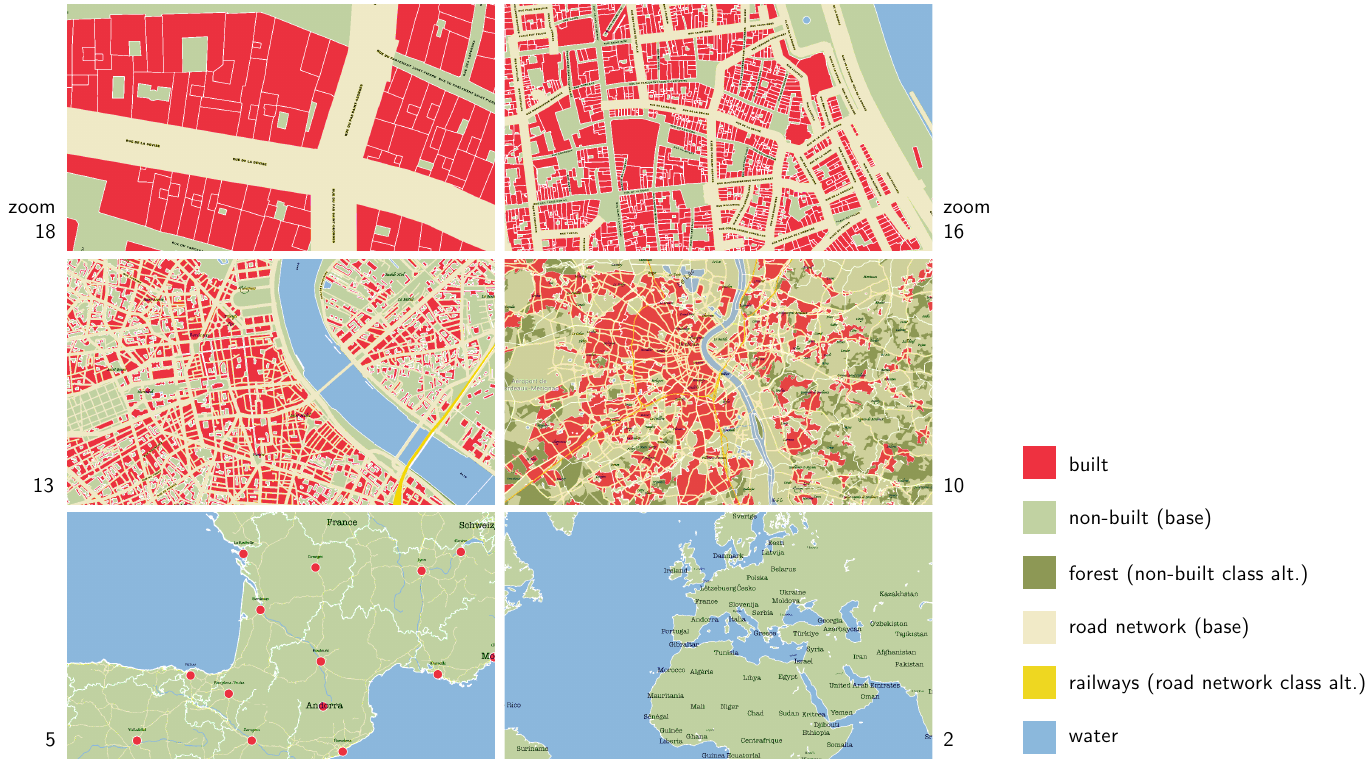}
\caption{Semantic masks retrieved from MapTiler API, displayed as a function of the zoom level. The legend distinguishes base components and alternative components (alt) of each subsequent segmentation class: forest is included in “non-built,” and railways in “road network” for processing purposes.}
\label{fig:2}
\end{figure}

In addition to manually annotated data, the training set includes synthetically generated data. Here, synthetic images are derived from MapTiler Planet \cite{2YTNGM73} contemporary reference geodata (Fig.~\ref{fig:2}). For this purpose, the spatial coverage and scale distributions of ADHOC are used as targets to constrain geodata sampling. The map scale $s$, expressed as a fraction (e.g., 1:20,000), is converted to an approximate zoom level $z$ by means of the following formula: $z\approx \alpha -{log}_{2}(\frac{1}{s})\ \ \ z\in {\mathbb{Z}}^{+},\ \ \alpha =28.1$. The resulting continuous estimate is rounded to an available nonnegative integer zoom level. Here, ${\mathbb{Z}}^{+}$ denotes the nonnegative integers, including zero. This conversion selects the MapTiler level of geographic detail and approximate feature size that best matches the nominal scale of each archival map.\footnote{MapTiler uses 512 × 512-pixel tiles and reports nominal scale at 96 PPI. At zoom level z, the equatorial Web Mercator resolution is $C/(512\cdot {2}^{z})$ metres per pixel, where C ≈ 40,075,016.7 m is Earth's equatorial circumference. Converting this resolution to a scale denominator gives $N\ \approx {2}^{28.1-z}$, hence $z\ \approx 28.1-{log}_{2}\left(N\right)=28.1-{log}_{2}\left(1/s\right)$, with $z$ constrained to an available integer zoom level. See MapTiler, “Map resolution and zoom,” \url{https://docs.maptiler.com/guides/how-maps-work/map-resolution/} (accessed 11 August 2026). Because Web Mercator ground resolution varies with latitude, the relation is used here as an approximation. The 96-PPI convention defines the zoom-to-scale relationship and is distinct from the 300-dpi output resolution assigned to the generated raster samples.} The geographic bounding box is calculated as a function of the zoom level and geographic coordinates, with a target dimension of 768\,×\,768 pixels, using WGS84 pseudo-Mercator projection (EPSG:3857) and a resolution of 300 dpi.

The complete list of object classes queried is provided in Table~\ref{tab:a4}, in Appendix. The objective is not to reproduce historical data with factual precision, but rather to approximate their visual appearance. For instance, airport names constitute anachronistic data points; nonetheless, they may be useful for labeling larger cities with a distinctive typography. Certain features are also restricted to specific zoom levels. Additionally, certain object classes, like forests, were randomly hidden to reflect cartographic incompleteness. The selection of contemporary features was refined iteratively through qualitative assessments informed by expertise in the content and stylistic conventions observed in historical maps. 

\begin{figure}[pos=htbp,align=\centering]
\includegraphics[width=.95\linewidth,height=.90\textheight,keepaspectratio]{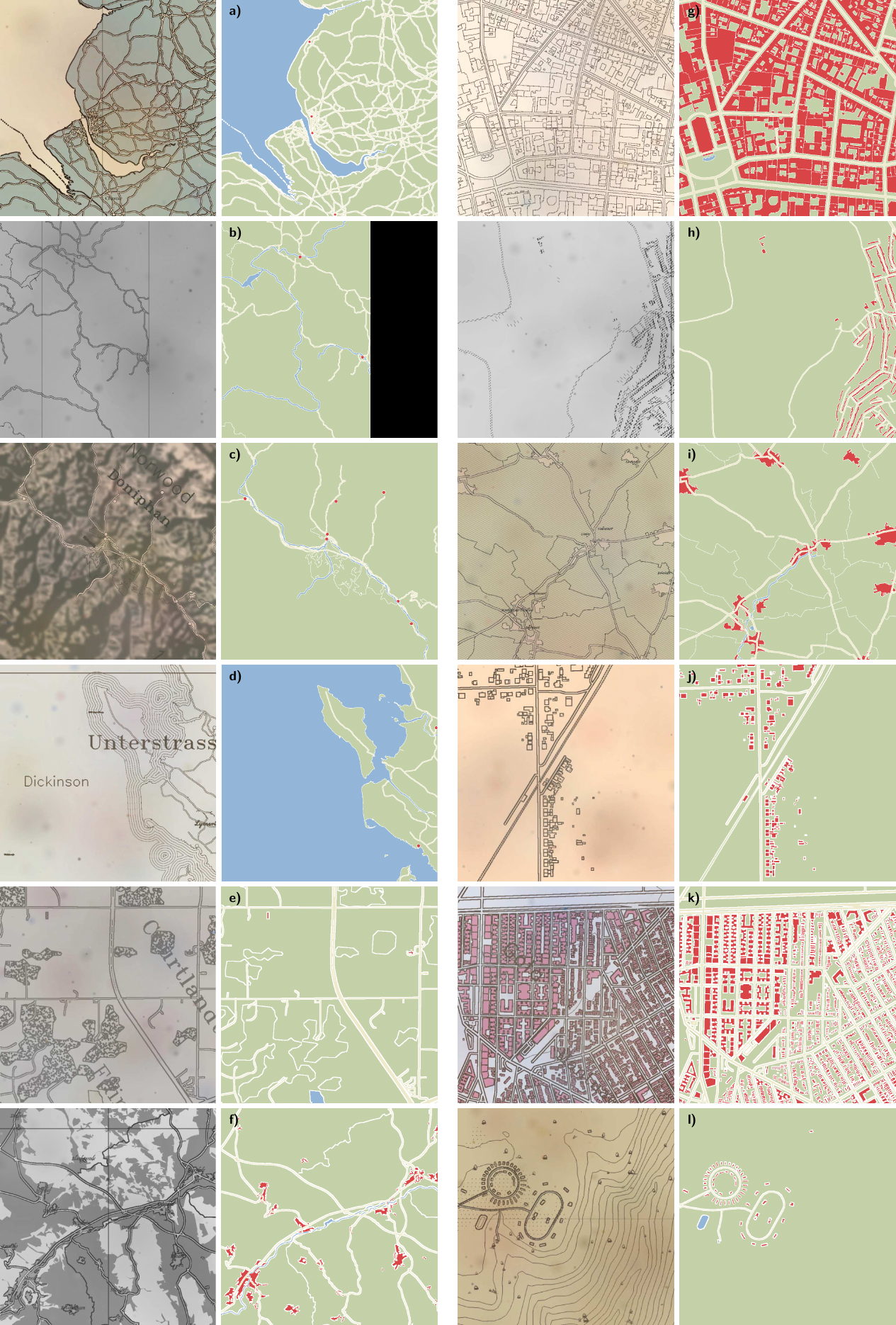}
\caption{Synthetically generated training samples. Synthetic map images (left) and corresponding annotations (right). The annotation color code follows the legend of Fig.~\ref{fig:2}.}
\label{fig:3}
\end{figure}

In total, 12,122 raster samples were collected using the MapTiler Application Programming Interface (API). A stylization process was subsequently employed to generate visually plausible synthetic maps (Fig.~\ref{fig:3}). Stylization entails the probabilistic application of graphical processes to the template samples. Here, a variety of generative drawing functions were used to produce of diverse synthetic maps. Parameters of the stylization algorithm—e.g., color distributions and the relative frequency of graphical processes—were initially derived from annotated Semap data and iteratively adjusted based on qualitative assessments. The graphical processes implemented include plain color (e.g., Fig.~\ref{fig:3}a, 3k), dotted patterns (e.g., Fig.~\ref{fig:3}e, 3l), hatchings (e.g., Fig.~\ref{fig:3}h, 3i), waterlines (e.g., Fig.~\ref{fig:3}d), texture masks extracted from Semap, and icon sprites \cite[p.~207]{PJC6ZP3W}, e.g., Fig.~\ref{fig:3}l). The color of each pattern is sampled according to the specific class, from a Gaussian mixture derived from Semap images. To increase color variability, darker spots were also applied to the canvas. Rescaling and anti-aliasing are employed to reproduce natural color interpolation. Fifteen percent of the images were converted to grayscale (e.g., Fig.~\ref{fig:3}b, 3f). In addition, images were saved in JPEG format to replicate compression artifacts.

Since relief is a recurrent feature of map images and can constitute a cue that might help recognize certain geographic contexts, the stylization algorithm was also used to approximate relief, using three common representation processes: hatching (e.g., Fig.~\ref{fig:3}h), hillshading (e.g., Fig.~\ref{fig:3}c), and isolines (e.g., Fig.~\ref{fig:3}l). Elevation data were retrieved through the Mapbox TerrainV2 API \cite{GYRSWYW7}. 

Finally, the image was randomly cropped along one edge or at a corner to imitate the map frame (e.g., Fig.~\ref{fig:3}b). The thicknesses of linear features (e.g., watercourses or the road network) and boundary lines were randomized to increase representational diversity. To prevent the segmentation model from overfitting to the relative positions of text labels and to enhance typographic diversity, additional text labels sampled from the repertoire of place names in the ADHOC database were randomly added (e.g., Fig.~\ref{fig:3}c–\ref{fig:3}e, 3k). Finally, an artificial graticule was occasionally imprinted on the image, in the form of horizontal or vertical lines that intersect to imitate a map grid (e.g., Fig.~\ref{fig:3}f).

\begin{figure}[pos=htbp,align=\centering]
\includegraphics[width=\linewidth]{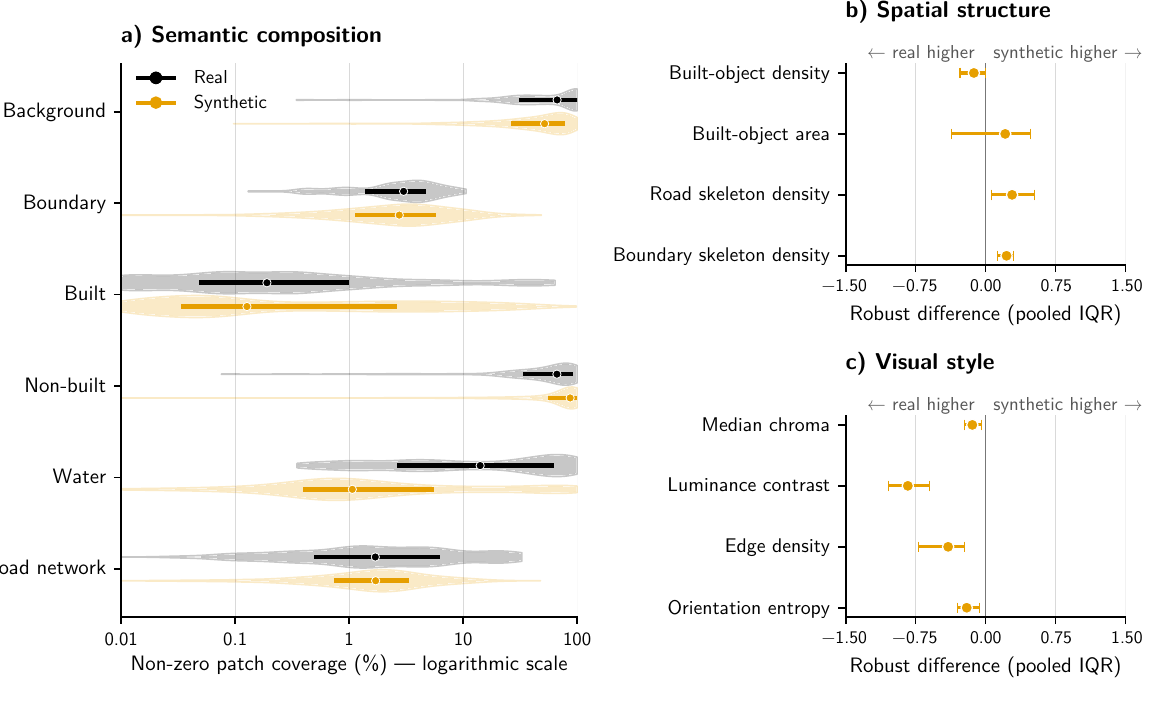}
\caption{Synthetic and real data characteristics. The synthetic corpus (n = 12,122) is compared with the real Semap test partition (n = 144). (a) Non-zero per-patch semantic-class coverage, conditional on class presence, on a logarithmic scale from 0.01\% to 100\%. Points and thick lines indicate medians and interquartile intervals. Shaded areas indicate the distribution shape. (b) Spatial-structure and (c) visual-style differences, both expressed as the synthetic-minus-real difference in medians divided by the pooled interquartile range. Positive values indicate larger values in synthetic data, and bars indicate patch-bootstrap 95\% confidence intervals. Both effect panels share the same horizontal scale from –1.5 to 1.5. Images and annotations were standardized to 768 × 768 pixels. Spatial measurements are conditional on presence of the corresponding semantic class.}
\label{fig:4}
\end{figure}

To quantify how closely the procedural data reproduce the test target domain, Figure~\ref{fig:4} compares the 12,122 synthetic samples with the 144 real images in the Semap test partition. Panel (a) reports distributions of non-zero per-patch semantic-class coverage. Panels (b) and (c) report robust synthetic-minus-real differences in medians, standardized by the pooled interquartile range. Bars show 95\% patch-bootstrap confidence intervals. Spatial measurements are conditioned on the presence of the corresponding class.

The two datasets show broad overlap for several geographic classes, although systematic differences remain. Water tends to cover a larger share of the real test patches. Historical map corpora are themselves known to under-represent maritime spaces, such as seas and oceans, relative to physical geography \cite[pp.~151--154]{PJC6ZP3W}. The difference observed here therefore indicates that water remains under-represented by the generator relative to this test subset. Spatial structure is realistic overall: the confidence intervals for built-object density and area include zero, while road and boundary skeleton densities are only moderately higher in synthetic annotations. The latter tendency may be useful during pretraining because roads and boundaries are among the most difficult classes to recognize.

The visual-style comparison captures quantitative differences in graphical properties explicitly targeted by the procedural generator. Median chroma is the median CIELAB distance from neutral grey and therefore measures typical colour saturation independently of lightness and hue. Its slightly lower value indicates that synthetic colours are somewhat more muted. Orientation entropy, which quantifies the diversity of strong gradient directions, is also modestly lower, consistent with more regular synthetic textures such as hatchings. Synthetic images also have lower luminance contrast and edge density. These properties may make pretraining more demanding and discourage reliance on colour or crisp edges alone, favoring morphological cues and contextual hierarchy; they nevertheless also reveal a remaining visual domain gap. The higher edge density in real images, despite their lower boundary prevalence, suggests that many edges occur within semantic surfaces through irregular colouring, texture, text, or printing rather than only at annotated class boundaries.

\FloatBarrier

\section{Segmentation approach}\label{sec:4}

\subsection{Choice of model architecture}\label{sec:4.1}

We rely on Mask2Former, a masked-attention Mask Transformer \cite{TQADKIQH} with a Swin-L backbone \cite{AGXDA7XV} for semantic segmentation. The choice of backbone is justified by the ability of Swin-Transformers to accommodate multiscale image objects, given their hierarchical design. The implementation used in this paper relies on MMSegmentation \cite{4F8ZKYWC}.

\subsection{Training approach}\label{sec:4.2}

The Semap dataset was split into three partitions (70/20/10\%) for training, validation, and testing, respectively. Synthetic data were employed exclusively for training (80\%) and validation (20\%). Under this configuration, they constitute 90.6\% of the training set. Conventional online augmentation comprises random 768 × 768 cropping and horizontal flipping. This is only one source of training variation: the procedural generator performs substantial domain randomization of colour, texture, line width, typography, relief depiction, grayscale rendering, map framing, and compression artefacts, as detailed in Section~\ref{sec:3.2}.

The model was trained for 345,000 iterations with a batch size of 1 and a multiscale composite loss. During the first 205,000 iterations, the model was trained on both Semap and synthetic data. It was then tuned on Semap only for the remaining 140,000 iterations. At each stage, only the model that achieved the highest validation performance was retained (early stopping policy).

For the HCMSSD benchmarks, three training regimes are reported. “Generic” denotes the generic Semap checkpoint evaluated directly on the target benchmark without target fine-tuning. “Transfer” denotes target-specific fine-tuning initialized from that generic checkpoint. “Retrained” denotes a separate Mask2Former–Swin-L model trained on the corresponding Paris or World target data from a generic ADE20K initialization, without Semap-specific pretraining.

The loss function combines binary cross-entropy (BCE) for mask prediction, cross-entropy (CE) for class attribution, and Dice loss \cite{JUK5Y27S} to mitigate class imbalance. A lower weight is assigned to the boundary class because its relative difficulty could otherwise hinder training at the expense of the land classes prioritized in this study. Training relies on the AdamW optimizer \cite{IT57HISL} with a polynomial-decay learning-rate schedule.

\subsection{Inference strategy \& hierarchical integration}\label{sec:4.3}

Since physical map documents are often quite large, the corresponding digital images can exceed 10,000\,×\,10,000, or even 20,000\,×\,20,000 pixels. Consequently, each image had to be partitioned into smaller 768\,×\,768 pixel patches—the dimension on which the model was trained. At inference, windows with 64-pixel overlaps were employed to mitigate border effects arising from the disruption of context at patch boundaries.

Moreover, while the selected architecture already favors hierarchical image modeling, map images are so fundamentally multiscale that inference is applied at two image scales and resolutions: first at the original resolution and then at half that resolution. The objective is to improve the recognition of large objects spanning several patches. Intuitively, this strategy may also enhance segmentation quality by enabling consensus predictions at a reduced computational cost, owing to the decreased spatial resolution. In the end, the raw logits from overlapping and multiscale prediction maps are averaged, and each pixel is assigned to the semantic class with the highest logit.

\subsection{Performance evaluation}\label{sec:4.4}

Let $N$ be the number of test images and $K=6$ the number of semantic classes. We write ${Y}_{i,k}$ the set of pixels $p$ in image $i$ that belongs to class $k$, and ${\hat{Y}}_{i,k}$ the set of pixels in image $i$ whose predicted class is $k$. We also define ${\cup }_{i,k}$= ${Y}_{i,k}\cup {\hat{Y}}_{i,k}$, the union of ${Y}_{i,k}$ and ${\hat{Y}}_{i,k}$, and ${\cap }_{i,k}$= ${Y}_{i,k}\cap {\hat{Y}}_{i,k}$ the intersection. For every class and image, we compute the precision, recall, and the intersection over union (IoU):

\[{Prec}_{i,k}=\frac{\sum _{\forall p} {\cap }_{i,k}}{\sum _{\forall p} {\hat{Y}}_{i,k}}\]

\[{Rec}_{i,k}=\frac{\sum _{\forall p} {\cap }_{i,k}}{\sum _{\forall p} {Y}_{i,k}}\]

\[{IoU}_{i,k}=\frac{\sum _{\forall p} {\cap }_{i,k}}{\sum _{\forall p} {\cup }_{i,k}}\]

For comparison, the arithmetic mean of mean precision and mean recall is reported as mPR; it should not be interpreted as the harmonic F1 score. The contribution of each sample image $i$ to the class averages is then normalized with respect to the ground-truth count $\sum _{\forall p} {Y}_{i,k}$. This procedure ensures, for example, that an image containing only a small house does not influence the built class average as much as a map predominantly depicting an urban environment. The stabilizing impact of this choice on heterogeneous datasets is demonstrated in Table~\ref{tab:a3} in the Appendix. For each metric ${M}_{k}\in \{{Prec}_{k},{Rec}_{k},{IoU}_{k}\}$, we thus have:

\[{M}_{k}=\frac{\sum _{i=1}^{N} {M}_{i,k}\sum _{\forall p} {Y}_{i,k}}{\sum _{i=1}^{N} \sum _{\forall p} {Y}_{i,k}}\]

Global Semap performance is calculated as the mean of the four geographic classes—built, non-built, water, and road network. For the HCMSSD–Paris and HCMSSD–World benchmarks, all reproduced and proposed-model predictions are evaluated using the standard per-class macro-average, also used by Arzoumanidis et al. \cite{D33PL94I}: counts are first accumulated across the complete validation benchmark for each foreground class, and the resulting four class scores are then averaged without class weighting (see Table~\ref{tab:a2}, Appendix).

\subsubsection{Assessment of cross-cultural genericity and performance biases}\label{sec:4.4.1}

To verify the assumption of genericity and to control for the presence of systematic performance biases, segmentation performance is evaluated against multiple metadata variables (collection institution, publication country, coverage country, map scale, publication year). Specifically, an ordinary least squares (OLS) model is fitted, using partition-z-standardized mean IoU (${mIoU}_{z}$) per patch as dependent variable. The categorical explanatory variables (noted $C\left(\cdot \right)$, collection institution, publication country, and coverage country) are one-hot encoded, and rare categorical values (n < 5) are excluded. The map-scale variable is treated logarithmically, and both continuous variables (map scale, publication year) are min-max normalized. The multivariate statistical model is given below.

\[\begin{aligned}&{mIoU}_{z}\ \sim \ C\left(institution\right)+C\left(pub.\ country\right)+C\left(cov.\ country\right)\\ &+{log\left(scale\right)}_{norm}+{pub.\ year}_{norm}\end{aligned}\]

\FloatBarrier

\section{Results of segmentation}\label{sec:5}

\subsection{Overall performance}\label{sec:5.1}

\subsubsection{Performance per class}\label{sec:5.1.1}

Table~\ref{tab:1} reports semantic segmentation performance on the Semap test set. The model performs best on the built and non-built classes, with IoU values of 79.8\% and 81.8\%, respectively. Water is also well segmented, achieving 72.2\% IoU. The least accurately recognized geographic class is the road network, at 62.9\% IoU. Boundaries, which were assigned a lower weight during training, exhibit 40.7\% IoU, whereas the map background is well segmented, at 76.8\%. Overall, the model shows high precision, ranging from 80.7\% to 92.4\% across geographic classes, with recall values between 72.7\% and 86.0\%. Mean performance across geographic classes reaches 74.2\% mIoU, 79.4\% recall, and 85.4\% precision.

\begin{table}[pos=htbp]
\centering
\footnotesize
\begin{adjustbox}{max width=\linewidth}
\begin{tabular}{@{}lrrrrrrrrr@{}}
\toprule
 & \multicolumn{3}{l}{Test set (base)} & \multicolumn{3}{l}{Test set (no multiscale)} & \multicolumn{3}{l}{Test set (no synth. pretrain)} \\
class & IoU & Recall & Precision & IoU & Recall & Precision & IoU & Recall & Precision \\
\midrule
background & 76.8 & 78.2 & 90.5 & 78.7 & 81.0 & 87.9 & 86.3 & 89.4 & 91.1 \\
boundary & 40.7 & 59.0 & 56.2 & 41.9 & 55.9 & 60.6 & 39.7 & 58.2 & 55.4 \\
built* & 79.8 & 84.4 & 92.4 & 70.8 & 75.5 & 89.4 & 74.6 & 84.1 & 87.7 \\
non-built* & 81.8 & 86.0 & 87.2 & 79.3 & 84.4 & 85.1 & 78.4 & 83.0 & 87.0 \\
water* & 72.2 & 74.4 & 81.2 & 68.4 & 70.6 & 73.6 & 67.0 & 68.6 & 80.3 \\
road network* & 62.9 & 72.7 & 80.7 & 62.0 & 71.4 & 77.8 & 56.3 & 65.7 & 81.7 \\
*mean & \textbf{74.2} & \textbf{79.4} & \textbf{85.4} & \textbf{70.1} & \textbf{75.5} & \textbf{81.5} & \textbf{69.1} & \textbf{75.4} & \textbf{84.2} \\
\bottomrule
\end{tabular}

\end{adjustbox}
\caption{Performance of semantic segmentation on the Semap dataset. The test partition contains 144 samples. IoU denotes intersection over union; the mean is computed across the four geographic classes marked with an asterisk.}
\label{tab:1}
\end{table}

\begin{table}[pos=htbp]
\centering
\footnotesize
\begin{adjustbox}{max width=\linewidth}
\begin{tabular}{@{}p{.19\linewidth}p{.36\linewidth}rrrr@{}}
\toprule
Reference & Model (training) & mIoU & mR & mP & mPR \\
\midrule
\cite{Polak} & \textit{TransUNet (Reproduced)} & 67.9 & 79.4 & 82.1 & 80.7 \\
\cite{U4F2XWXA} & \textit{UNet–ResNet101 (Reproduced)} & 60.7 & 74.7 & 75.1 & 74.9 \\
\cite{D33PL94I} & SCGCN–ResNet101 & 63.5 & 82.7 & 74.2 & 78.5 \\
\cite{QWHQBJDN} & \textit{HRNet–OCRNet (Reproduced)} & 69.1 & 80.3 & 82.8 & 81.5 \\
Ours & Mask2Former–Swin-L (generic) & 71.5 & 77.9 & \textbf{89.1} & 83.5 \\
Ours & Mask2Former–Swin-L (transfer) & \textbf{76.0} & 83.8 & 88.2 & 86.0 \\
Ours & Mask2Former–Swin-L (retrained) & \textbf{76.0} & \textbf{84.2} & 88.6 & \textbf{86.4} \\
\bottomrule
\end{tabular}

\end{adjustbox}
\caption{Performance benchmark on HCMSSD–Paris held-out validation set. The target split contains 262 training and 65 validation samples. All reproduced rows use the same five-class label space, 768 × 768 crop convention, deterministic seed (0), no multiscale inference, and dataset-level per-class macro-average over the four foreground classes, following the formula used by Arzoumanidis et al. \cite{D33PL94I}. Architecture-appropriate initialization and optimization are retained. Reproduced checkpoints are selected by validation mIoU. Each row represents one run. mR = mean recall, mP = mean precision, and mPR = (mP + mR)/2 (arithmetic mean, not harmonic F1). Training-regime labels for our model are defined in Section~\ref{sec:4.2}.}
\label{tab:2}
\end{table}

\begin{table}[pos=htbp]
\centering
\footnotesize
\begin{adjustbox}{max width=\linewidth}
\begin{tabular}{@{}p{.19\linewidth}p{.36\linewidth}rrrr@{}}
\toprule
Reference & Model (training) & mIoU & mR & mP & mPR \\
\midrule
\cite{Polak} & \textit{TransUNet (Reproduced)} & 60.3 & 74.3 & 75.9 & 75.1 \\
\cite{U4F2XWXA} & \textit{UNet–ResNet101 (Reproduced)} & 48.2 & 65.7 & 63.4 & 64.5 \\
\cite{QWHQBJDN} & \textit{HRNet–OCRNet (Reproduced)} & 59.0 & 73.2 & 75.4 & 74.3 \\
Ours & Mask2Former–Swin-L (generic) & \textbf{74.4} & \textbf{81.7} & \textbf{88.6} & \textbf{85.2} \\
Ours & Mask2Former–Swin-L (transfer) & \textbf{76.3} & \textbf{88.4} & \textbf{92.6} & \textbf{90.5} \\
Ours & Mask2Former–Swin-L (retrained) & 74.2 & 86.8 & 84.1 & 85.4 \\
\bottomrule
\end{tabular}

\end{adjustbox}
\caption{Performance benchmark on HCMSSD–World held-out validation set. The target split contains 244 training and 61 validation samples. All reproduced rows use the same five-class label space, 768 × 768 crop convention, deterministic seed (0), no multiscale inference, and dataset-level per-class macro-average over the four foreground classes, following the formula used by Arzoumanidis et al. \cite{D33PL94I}. Architecture-appropriate initialization and optimization are retained. Reproduced checkpoints are selected by validation mIoU. Each row represents one run. mR = mean recall, mP = mean precision, and mPR = (mP + mR)/2 (arithmetic mean, not harmonic F1). Training-regime labels for our model are defined in Section~\ref{sec:4.2}.}
\label{tab:3}
\end{table}

\subsubsection{Comparison benchmark}\label{sec:5.1.2}

Tables~\ref{tab:2} and \ref{tab:3} show that Mask2Former–Swin-L achieves the strongest performance on both HCMSSD–Paris and HCMSSD–World benchmarks. It achieves markedly higher scores than the reproduced UNet–ResNet101, TransUNet, and HRNet–OCRNet architectures. On Paris, the best Mask2Former–Swin-L result reaches 76.0\% mIoU, 6.9 percentage points (pp) above the strongest reproduced baseline, HRNet–OCRNet (69.1\%). On World, it reaches 76.3\% mIoU, 16.0 pp above the strongest reproduced baseline, TransUNet (60.3\%). The model also exceeds the SCGCN result reported for Paris (63.5\% mIoU). The results illustrate the genericity of the model. Whereas the reproduced baselines lose 7.6–12.5 pp between Paris and the more diverse World benchmark, Mask2Former–Swin-L maintains similar performance across both datasets. The “generic” configuration, which applies the generic Semap-and-synthetic checkpoint without target-specific fine-tuning, remains strong, especially on World (74.4\% mIoU). The “transfer” configuration reaches 76.3\% on World, compared with 74.2\% for target-specific retraining, indicating that heterogeneous pretraining is particularly useful for generic recognition. On the geographically coherent Paris benchmark, transfer and target-specific retraining both reach 76.0\%, showing that dataset-specific optimization can still capture geographic and stylistic specificities.

\subsubsection{Model capacity}\label{sec:5.1.3}

Table~\ref{tab:4} presents the performance of the model on the training, validation, and test partitions of Semap and reports the ablation study. The model attains 94.4\% mIoU on the training set, 75.5\% on the validation set, and 74.2\% on the test set. This disparity indicates that the model overperforms on the training data, signaling overfitting. At the same time, the excellent training fit demonstrates that optimization is effective and that the model has sufficient capacity for the task. This validates the choice of architecture and suggests that additional training data could lead to further performance improvements. Importantly, the close validation and test results provide no evidence of overfitting to the validation partition. Before real-data fine-tuning, validation mIoU is 70.9\%, whereas test mIoU is 69.1\%, a marginal difference. After fine-tuning, validation mIoU increases to 75.5\% and test mIoU reaches 74.2\%. These results confirm that real-data fine-tuning improves performance on unseen test samples.

\subsubsection{Impact of multiscale integration and synthetic data pretraining}\label{sec:5.1.4}

As shown in Table~\ref{tab:4}, multiscale integration and synthetic data pretraining both enhance segmentation performance. Ablation indicates that removing either one decreases mIoU by 4 to 5 percentage points. Synthetic data pretraining appears to benefit recall primarily, whereas multiscale integration improves both recall and precision to comparable extents. The impact on overall pixel-level accuracy, however, is more equivocal. As detailed in Table~\ref{tab:1}, this ambiguity is mostly attributable to the background class, which is better recognized without multiscale integration or synthetic data pretraining. Conversely, Table~\ref{tab:1} suggests that both strategies clearly improve the recognition of all four geographic classes, which constitutes the primary objective of the model. Multiscale integration seems to have a particularly positive effect on the recognition of the built (+9.0 IoU, +8.9 Recall, +3.0 Precision) and water classes (+3.8 IoU, +3.8 R, +7.6 P). The impact of synthetic data pretraining seems to be more distributed, with a significant impact on the recognition of the road network (+6.6 IoU, +7.0 R, –1.0 P), water (+5.2 IoU, +5.8 R, –0.9 P), built (+5.2 IoU, +0.3 R, +4.7 P), and non-built (+3.4 IoU, +3.0 R, +0.2 P) classes.

\begin{table}[pos=htbp]
\centering
\footnotesize
\begin{adjustbox}{max width=\linewidth}
\begin{tabular}{@{}lrrrrr@{}}
\toprule
Set/treatment & mIoU & mR & mP & mPR & Acc \\
\midrule
Test set (base) & \textbf{74.2} & \textbf{79.4} & \textbf{85.4} & \textbf{82.4} & \textbf{80.2} \\
– multiscale & 70.0 & 75.5 & 81.5 & 78.5 & 79.6 \\
– synth. pretrain. & 69.1 & 75.4 & 84.2 & 79.8 & \textbf{81.5} \\
– finetuning & 69.1 & 73.3 & 83.7 & 78.5 & 81.4 \\
Validation set & \textbf{75.5} & \textbf{80.5} & \textbf{87.8} & \textbf{84.1} & \textbf{84.5} \\
– finetuning & 70.9 & 76.3 & 86.7 & 81.5 & 82.7 \\
Training set & 94.4 & 96.9 & 97.1 & 97.0 & 97.8 \\
\bottomrule
\end{tabular}

\end{adjustbox}
\caption{Ablation study and comparison of test, validation, and training performance. The partitions contain 144 test, 289 validation, and 1,006 training samples. Treatments remove one feature at a time. mIoU = mean intersection over union, mR = mean recall, mP = mean precision, Acc = overall pixel accuracy, and mPR = (mP + mR)/2 (arithmetic mean, not harmonic F1). Means are computed over built, non-built, water, and road network.}
\label{tab:4}
\end{table}

\subsubsection{Confusion rates}\label{sec:5.1.5}

\begin{figure}[pos=htbp,align=\centering]
\includegraphics[width=.95\linewidth]{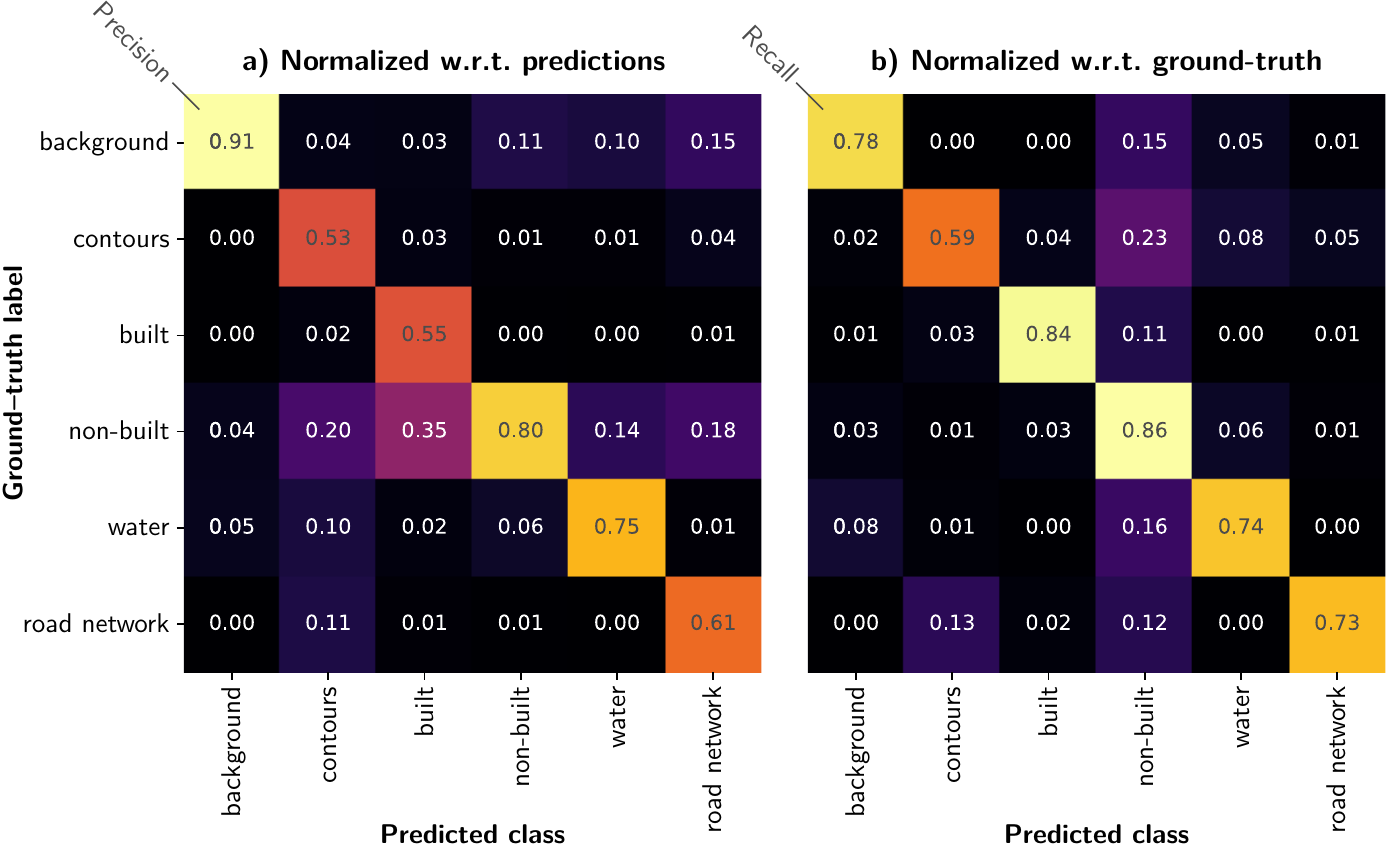}
\caption{Confusion matrix based on Semap test set predictions. The matrix is computed as the mean of the per-sample confusion matrices and normalized per class: (a) with respect to predictions and (b) with respect to ground truth. The diagonal values correspond to precision in (a) and recall in (b).}
\label{fig:5}
\end{figure}

Figure~\ref{fig:5} presents the confusion rates in the form of a confusion matrix. The diagonal of Figure~\ref{fig:5}a corresponds to precision values\footnote{These values slightly differ from those shown in Table~\ref{tab:1}. Here, the mean per-image confusion—that is, the average confusion over test samples—had to be used instead of the sample-normalized macro average (cf. Tab.~\ref{tab:a2}, and Tab.~\ref{tab:a3}). Therefore, each confused pixel is weighted equally regardless of the class to which it belongs.}. The results indicate that the boundaries, built, and road network classes appear to be overpredicted, mainly at the expense of the non-built class. This imbalance does not substantially affect the non-built class, however, since its area is comparatively larger (see Figure~\ref{fig:4}a). Some regions that should be classified as background are instead attributed to the non-built, water, or road network classes. Finally, portions of the water and road network classes tend to be misclassified as boundaries. This may correspond to linear objects, like waterways and thin roads.

\subsubsection{Assessment of cross-cultural genericity and performance biases}\label{sec:5.1.6}

\begin{figure}[pos=htbp,align=\centering]
\includegraphics[width=.95\linewidth]{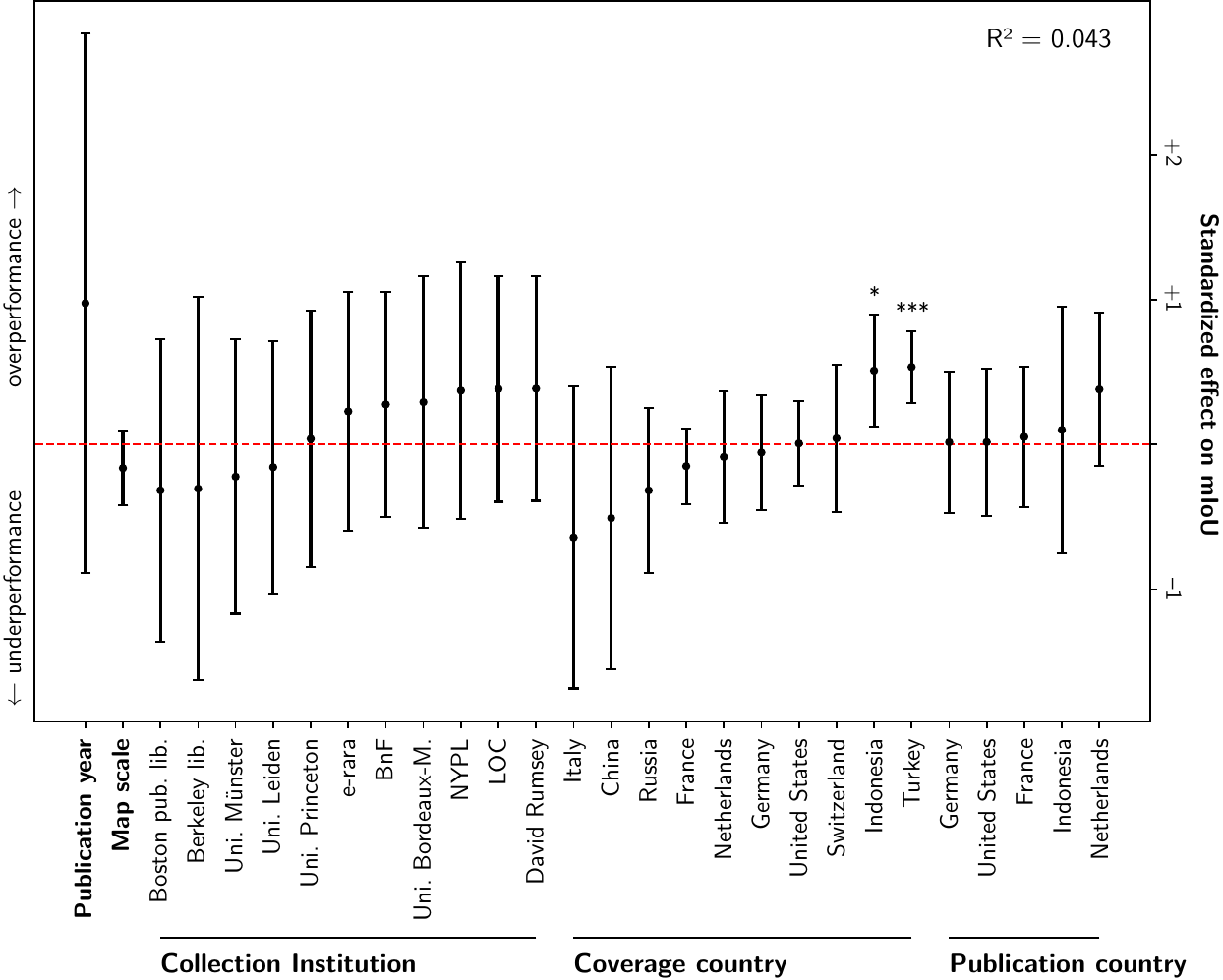}
\caption{Standardized effect of map-metadata variables on mIoU. Vertical bars indicate multivariate linear ordinary least-squares regression coefficients, with 95\% CI. The dependent variable, mIoU, denotes the per-patch (n=801) average intersection over union standardized within each dataset partition. No systematic bias is detected, apart from a slight positive effect for maps covering locations in Indonesia or Turkey. Overall, the regression explains little variance (R2 = 0.04), suggesting stable segmentation performance across evaluated metadata classes.}
\label{fig:6}
\end{figure}

Although the multivariate OLS model is statistically significant overall (**** p-value < 8·10\textsuperscript{-9}), it only explains a small fraction of the observed variance (R\textsuperscript{2} = 0.043). This suggests that semantic segmentation performance is largely consistent across tested metadata classes. Thus, no major systematic bias or failure mode is detected. Predictor coefficients are visualized in Figure~\ref{fig:6}. A slight overperformance is nevertheless observed for maps covering locations in Indonesia (+0.54 standard deviation, * p-value < 0.02) and Turkey (+0.51 std., *** p-value < 0.001). This particular outcome might reflect sampling errors and limited diversity within those sets derived from non-domestic map collections. More recent maps and larger-scale maps tend to be better segmented, yet this effect is not significant. Overall, the analysis shows minimal systematic influence of metadata characteristics on segmentation quality. The model seems to perform consistently across archival collections, geographic locations, scales, and historical periods, indicating robust potential for generalization.

\subsubsection{Calculation time and efficiency}\label{sec:5.1.7}

The segmentation model was trained on an older-generation (2014) Nvidia GeForce Titan X GPU equipped with 12 GB of VRAM. On this hardware, the network required 150 hours to converge. At inference time, the computation took 0.66 seconds per 768\,×\,768 image sample, which corresponds, on average, to 101 seconds per map, including multiscale integration and image reading/writing time.

\begin{figure}[pos=htbp,align=\centering]
\includegraphics[width=.95\linewidth,height=.90\textheight,keepaspectratio]{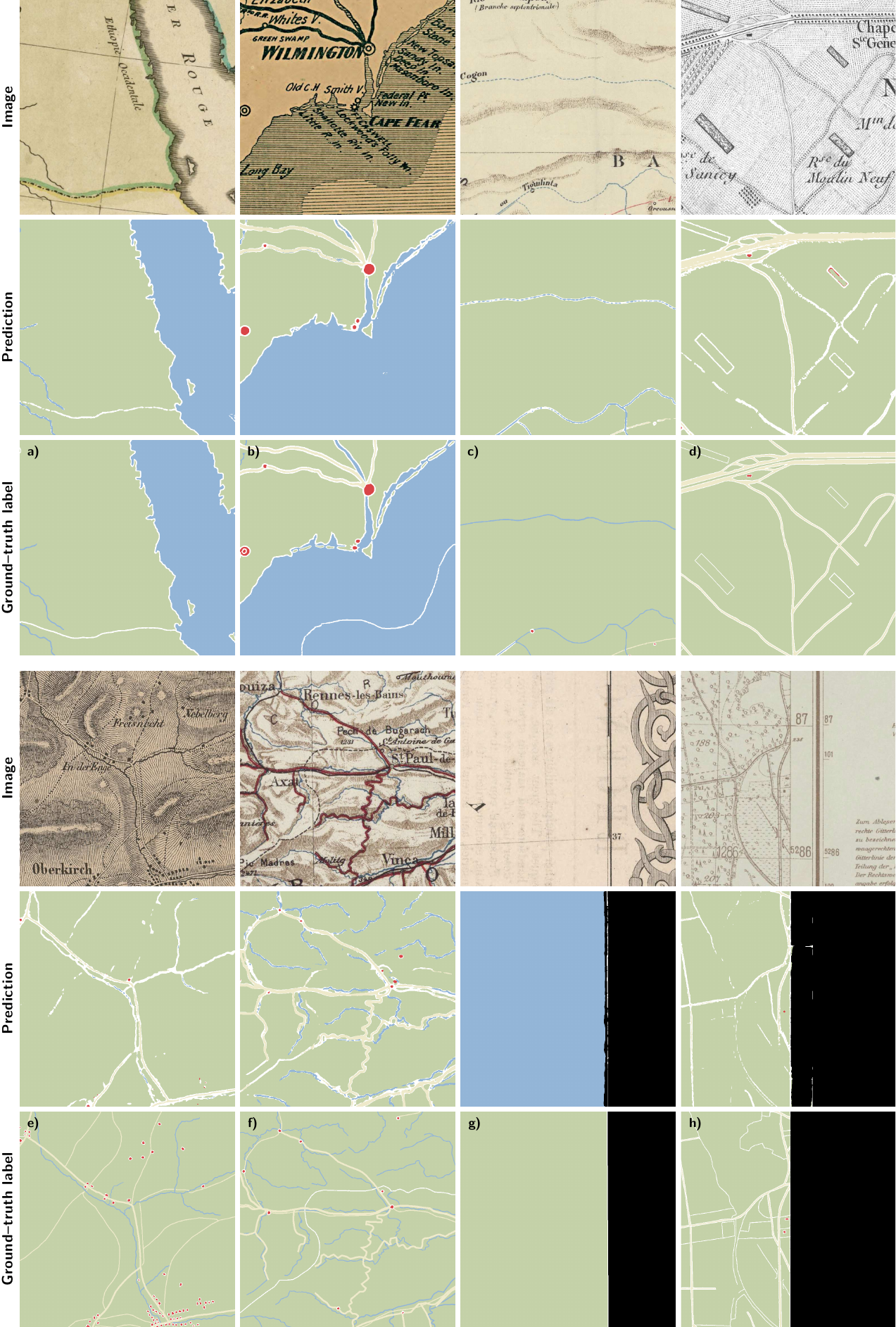}
\caption{Results of inference on the test set for a random subset of 8 samples.}
\label{fig:7}
\end{figure}

\subsection{Qualitative analysis}\label{sec:5.2}

\begin{figure}[pos=htbp,align=\centering]
\includegraphics[width=\linewidth,height=.43\textheight,keepaspectratio]{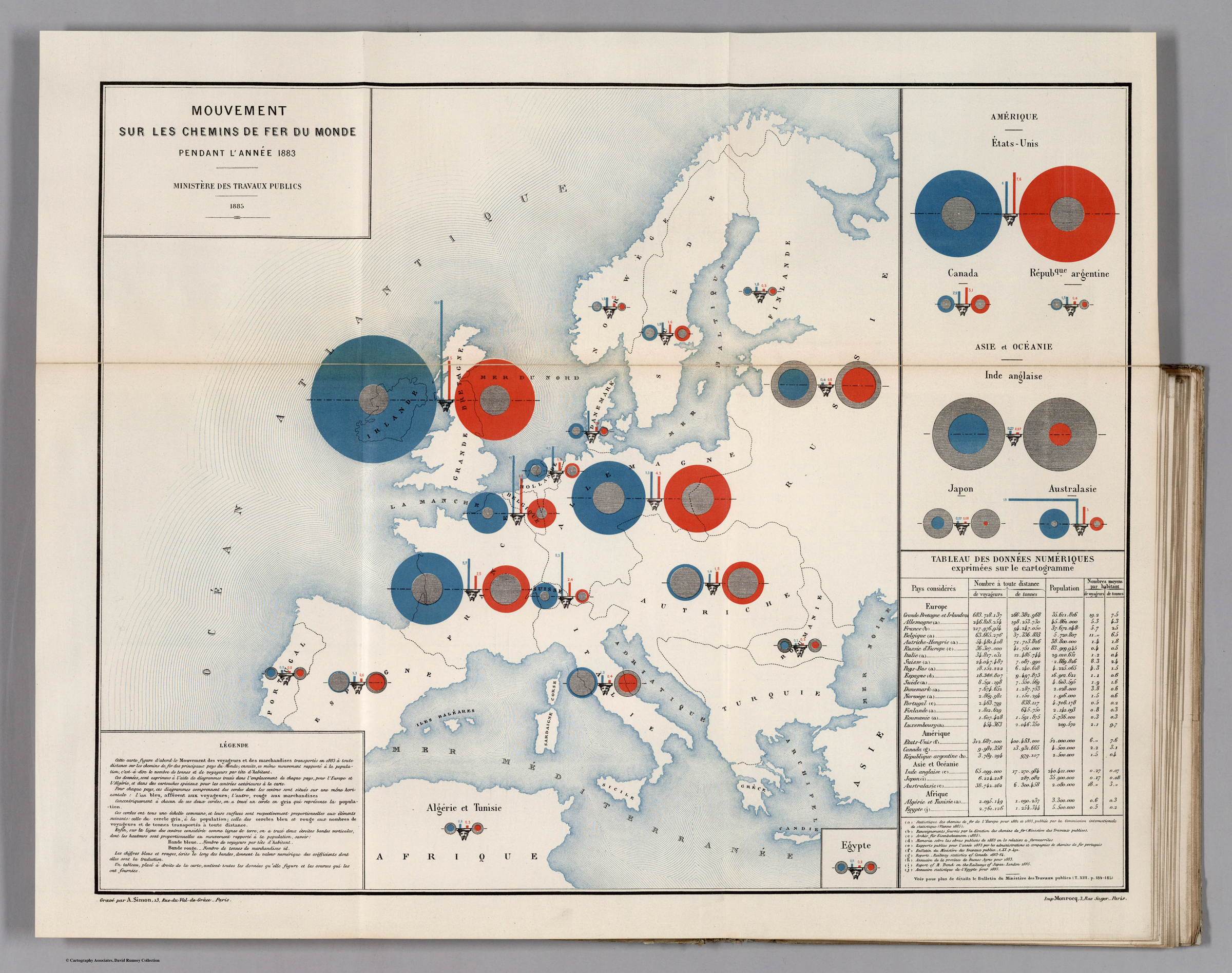}\\[.6em]
\includegraphics[width=\linewidth,height=.43\textheight,keepaspectratio]{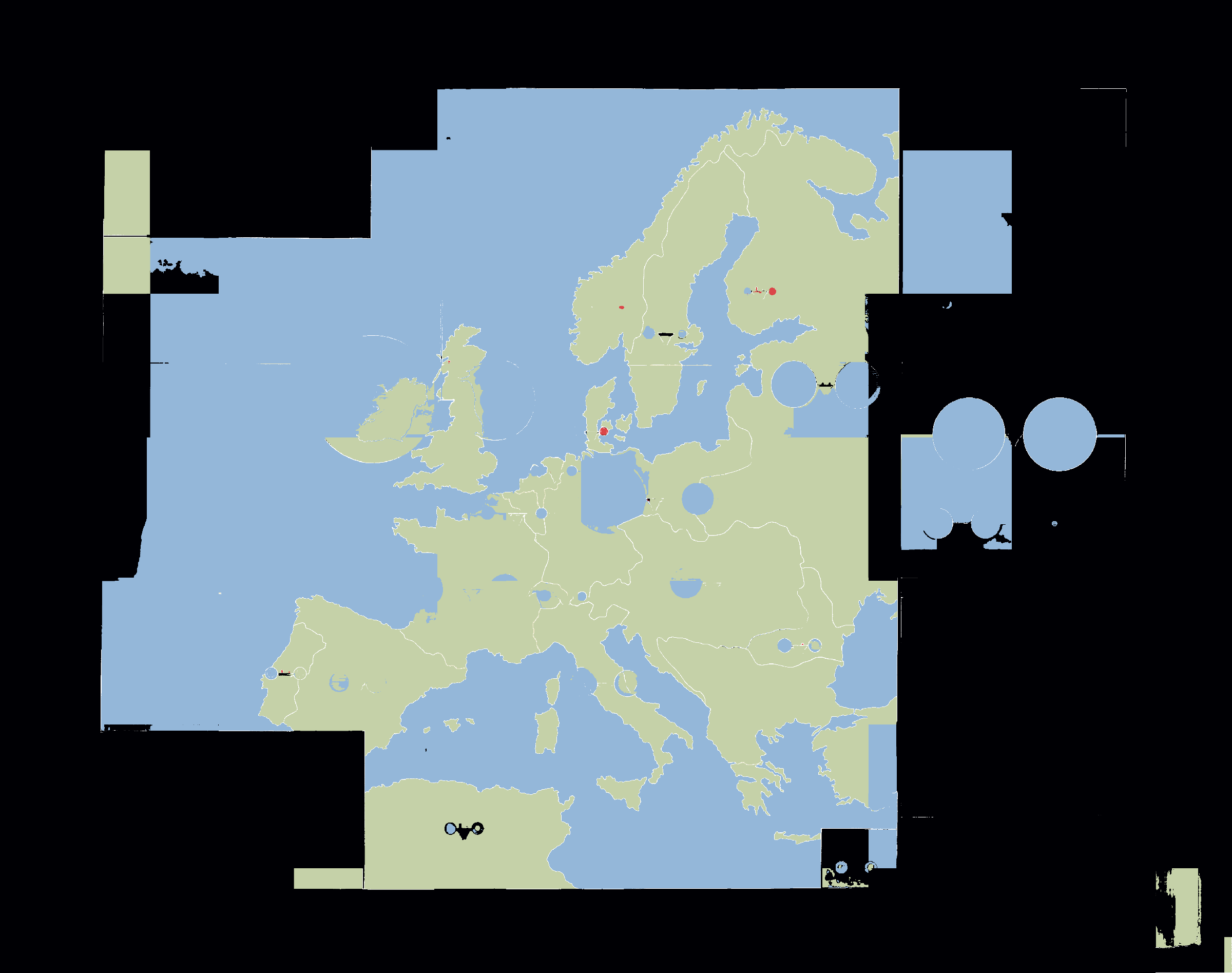}
\caption{(top) Map of railway trade volumes for 1883, published 1886. Émile Cheysson. Mouvement sur les chemins de fer du monde, 1883. Ministère des Travaux Publics, Paris. 49 x 60 cm. David Rumsey Collection, 12516.014. URL: \url{davidrumsey.com/luna/servlet/detail/RUMSEY~8~1~309240~90079149}; (bottom) Result of the semantic segmentation. The recognition of coastlines is impacted, although not entirely hindered by the presence of overlapping circles. Dark blue areas tend to be mistaken for water.}
\label{fig:8}
\end{figure}

\begin{figure}[pos=htbp,align=\centering]
\includegraphics[width=\linewidth,height=.43\textheight,keepaspectratio]{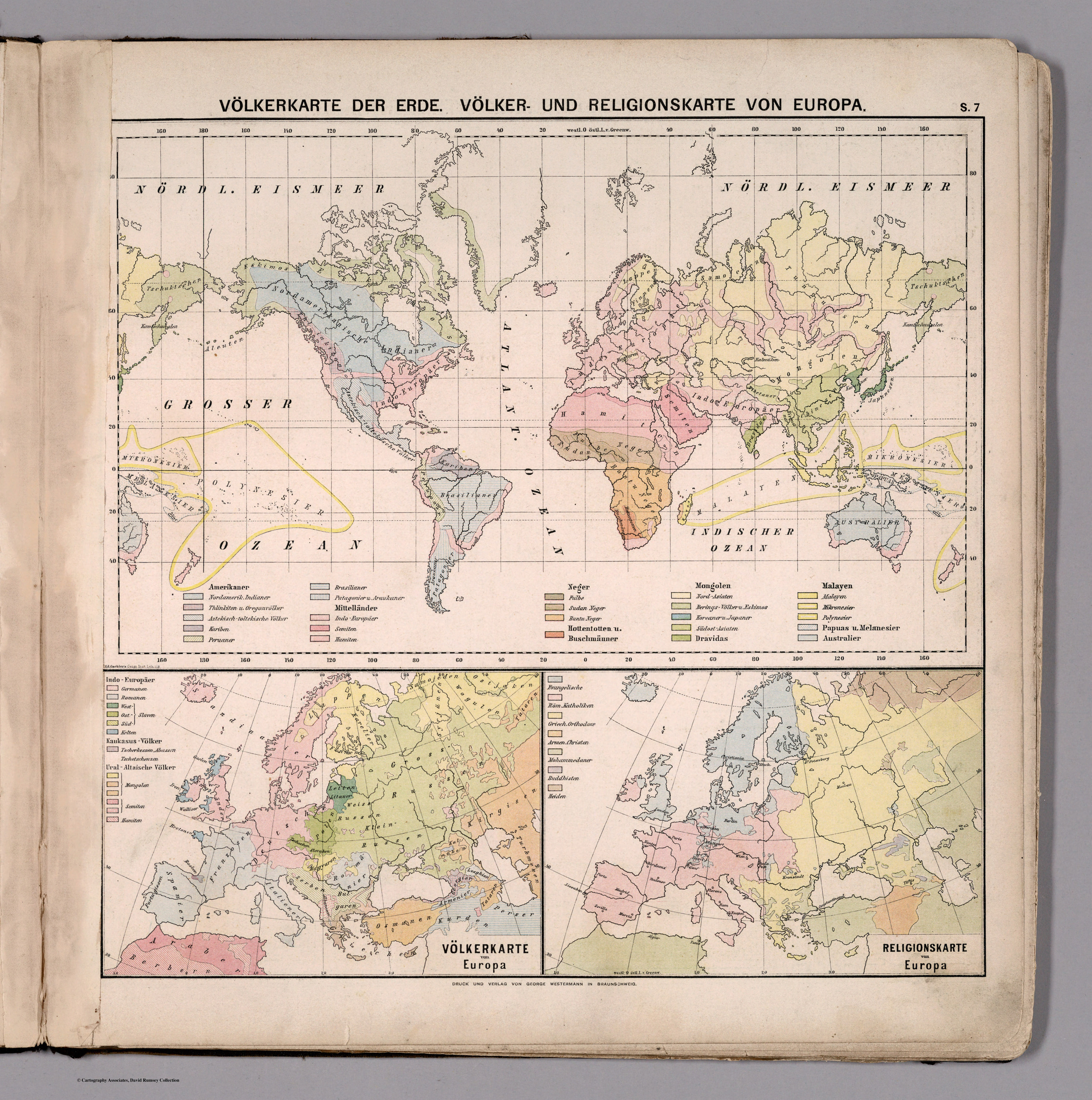}\\[.6em]
\includegraphics[width=\linewidth,height=.43\textheight,keepaspectratio]{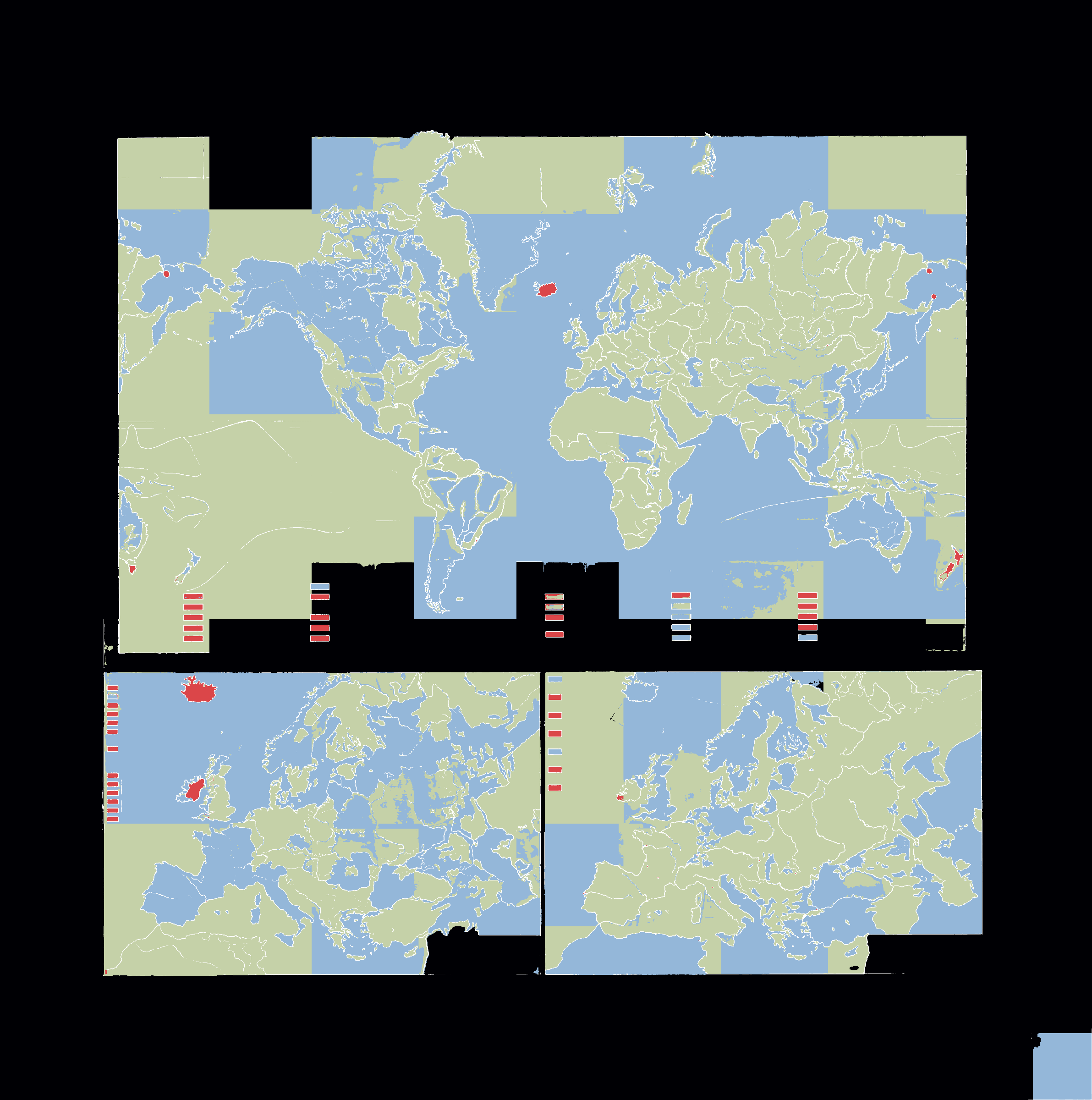}
\caption{(top) Ethnographic and religious map, 1883. Carl Diercke, Eduard Gaebler. Völkerkarte der Erde. Völker- und Religionskarte von Europa, 1883. Published by George Westermann, Braunschweig. 32 x 31 cm. David Rumsey Collection, 12198.011. URL: \url{davidrumsey.com/luna/servlet/detail/RUMSEY~8~1~311889~90081579}, (bottom) Result of the semantic segmentation. The distinction of land masses from seas seems affected by the use of a color code to represent thematic categories.}
\label{fig:9}
\end{figure}

\begin{figure}[pos=htbp,align=\centering]
\includegraphics[width=\linewidth,height=.43\textheight,keepaspectratio]{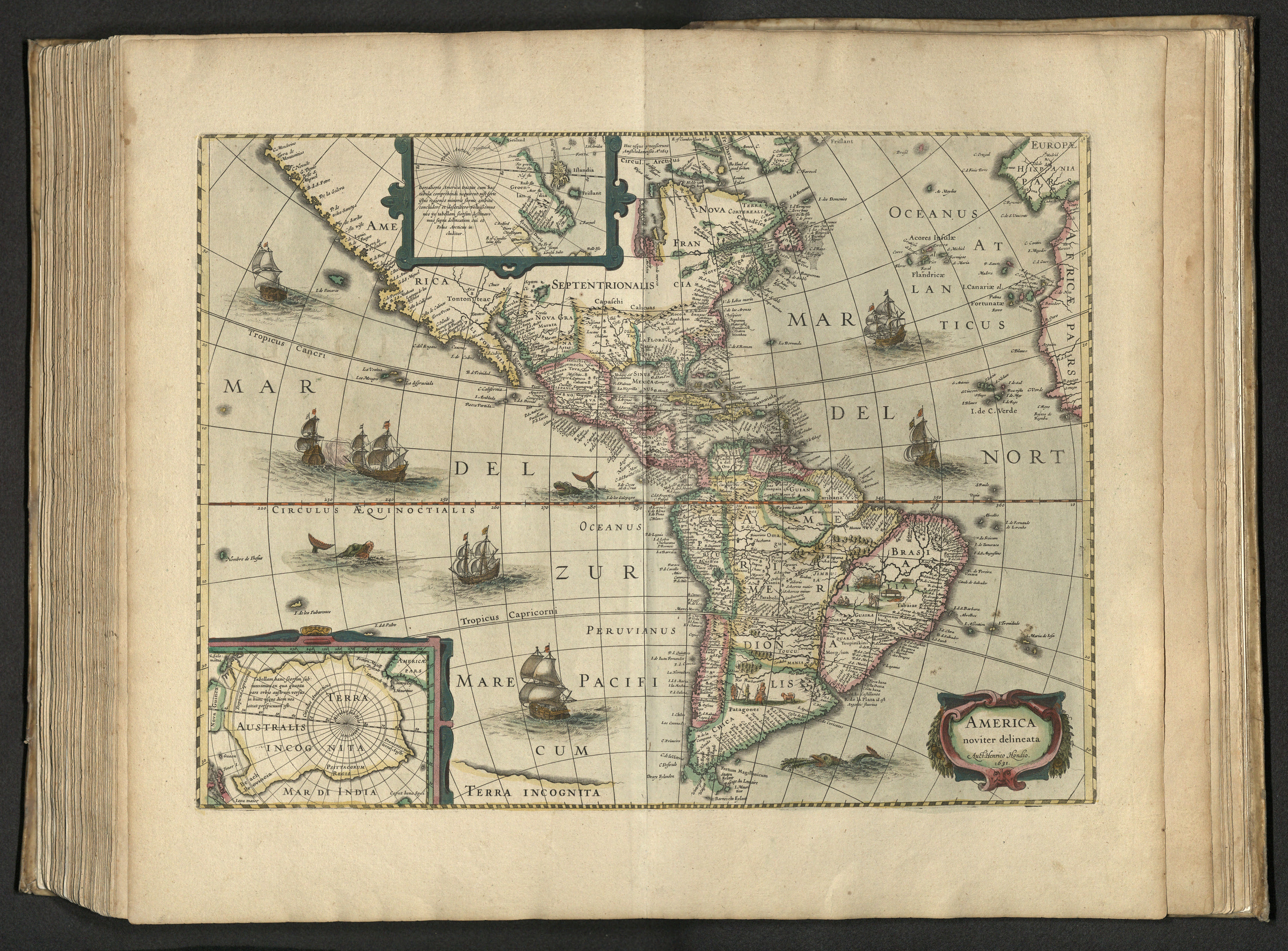}\\[.6em]
\includegraphics[width=\linewidth,height=.43\textheight,keepaspectratio]{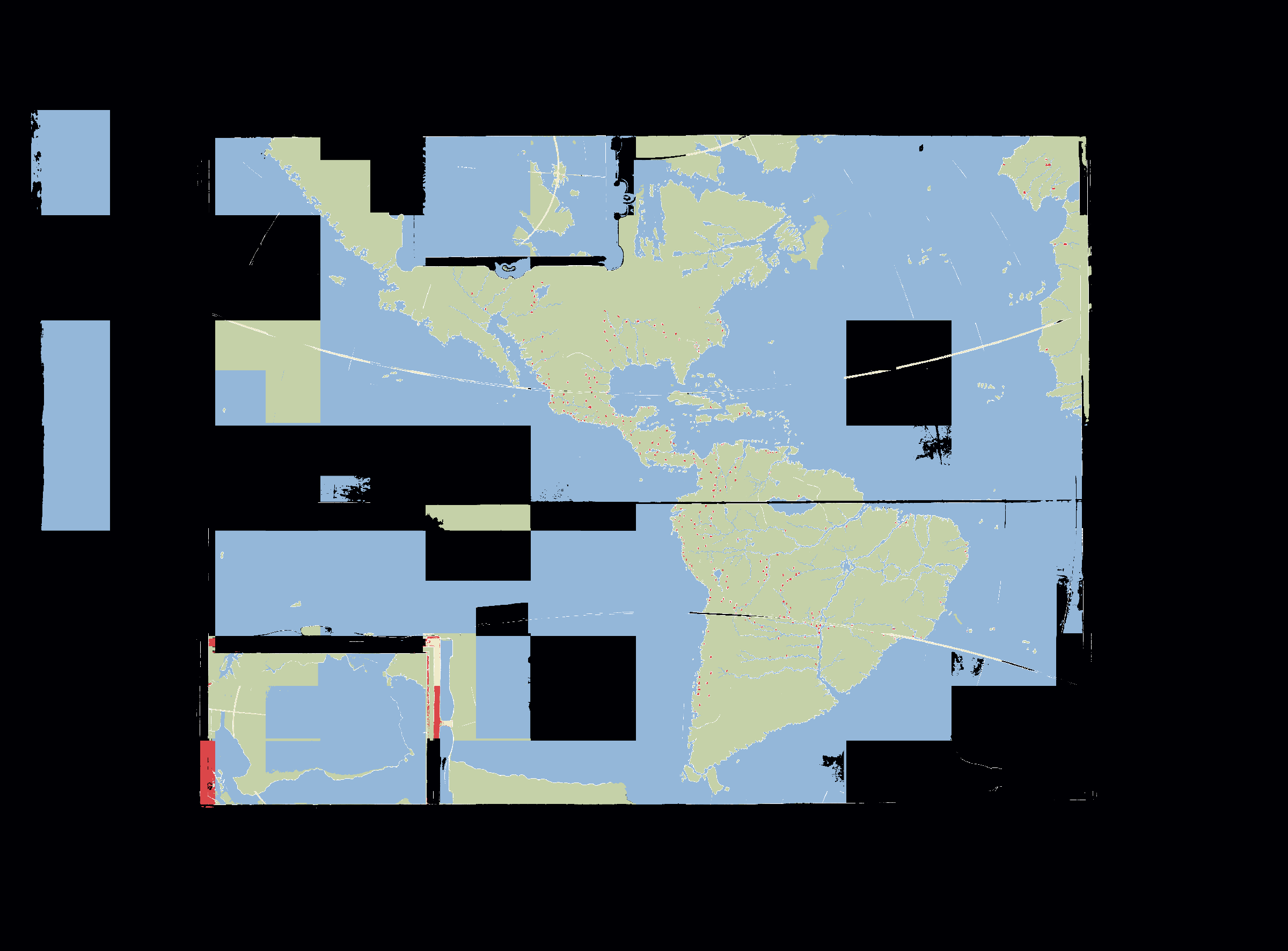}
\caption{(top) New map of America, part of the Atlas of Gerard Mercator and Jodocus Hondius, 1633. Hendrik Hondius. America noviter delineata, 1633. Published in Amsterdam. Copperplate, hand colored. 38 x 50 cm. David Rumsey Coll., 10621.195. URL: \url{davidrumsey.com/luna/servlet/detail/RUMSEY~8~1~345039~90107477}. (bottom) Result of the semantic segmentation. Landmasses are overall well segmented, whereas the recognition of seas seems impacted by the figuration of ship icons.}
\label{fig:10}
\end{figure}

\begin{figure}[pos=htbp,align=\centering]
\includegraphics[width=\linewidth,height=.43\textheight,keepaspectratio]{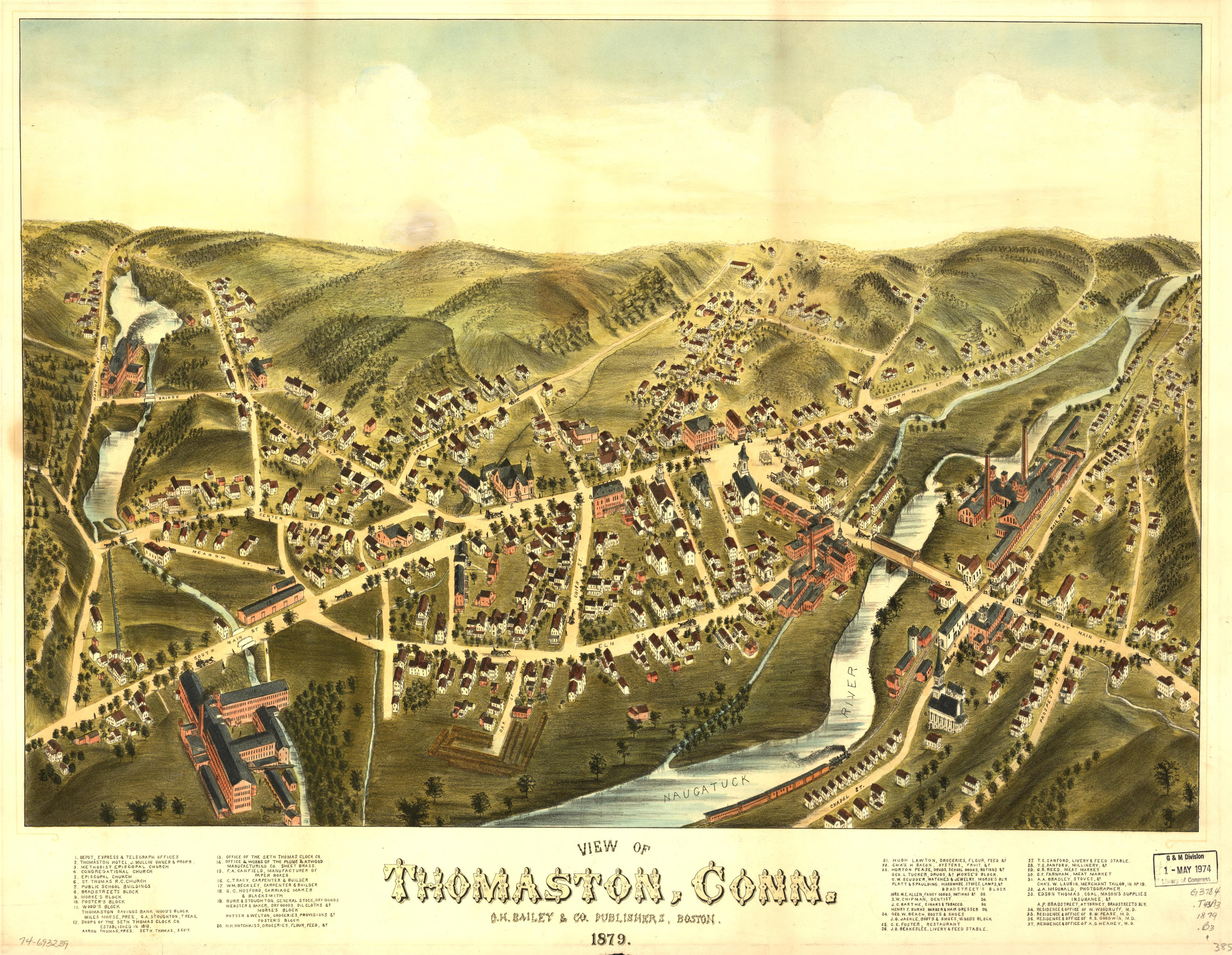}\\[.6em]
\includegraphics[width=\linewidth,height=.43\textheight,keepaspectratio]{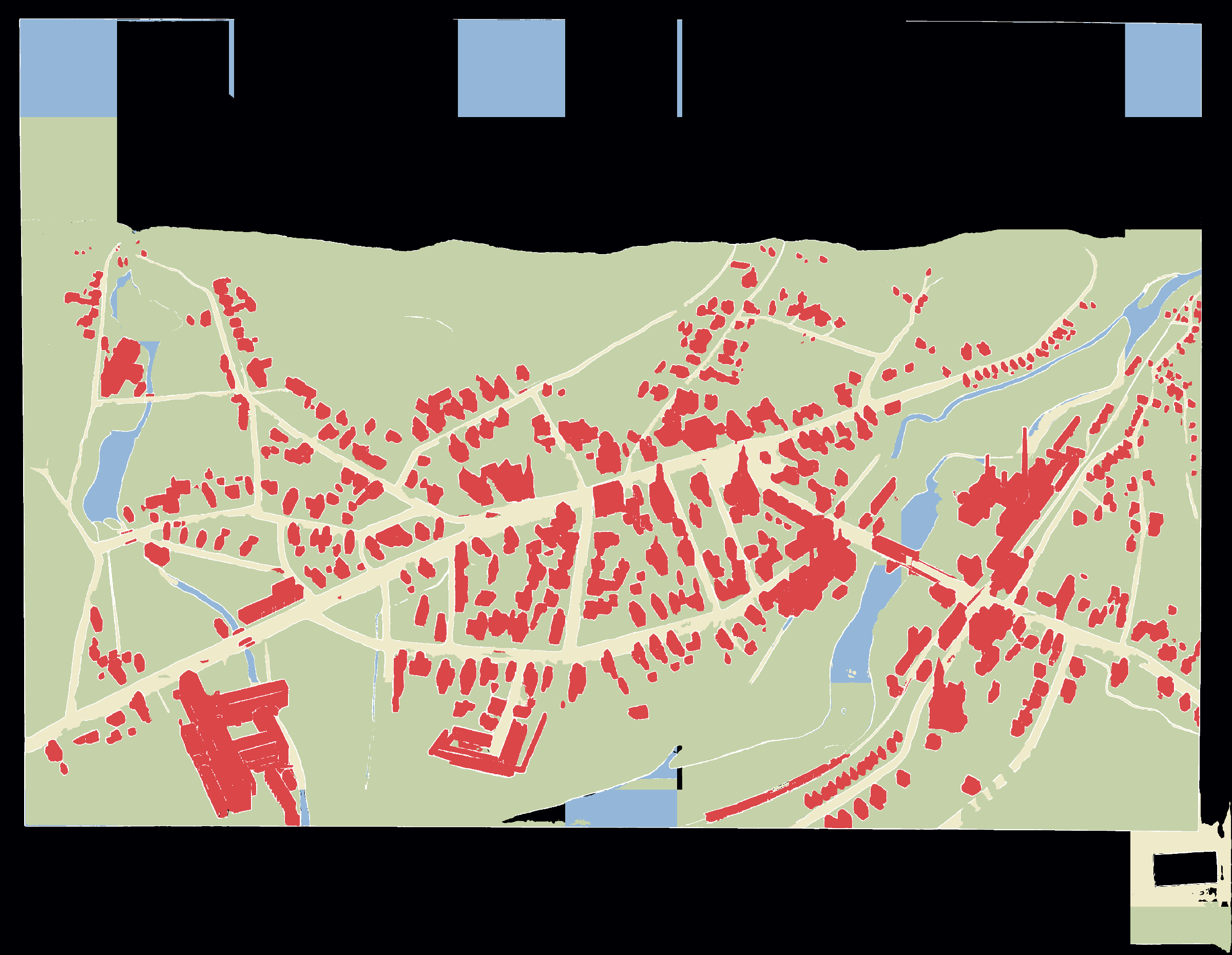}
\caption{(top) View of Thomaston, Connecticut, 1879. O.H. Bailey. View of Thomaston, Conn., 1879. O.H. Bailey \& Co, Boston. Hand colored. 43 x 63 cm. Library of Congress, G3784.T43A3 1879.B3. URL: \url{hdl.loc.gov/loc.gmd/g3784t.pm000974}. (bottom) Result of the semantic segmentation. Dwellings are well recognized despite the strong iconographic character of the representation.}
\label{fig:11}
\end{figure}

\begin{figure}[pos=htbp,align=\centering]
\includegraphics[width=\linewidth,height=.43\textheight,keepaspectratio]{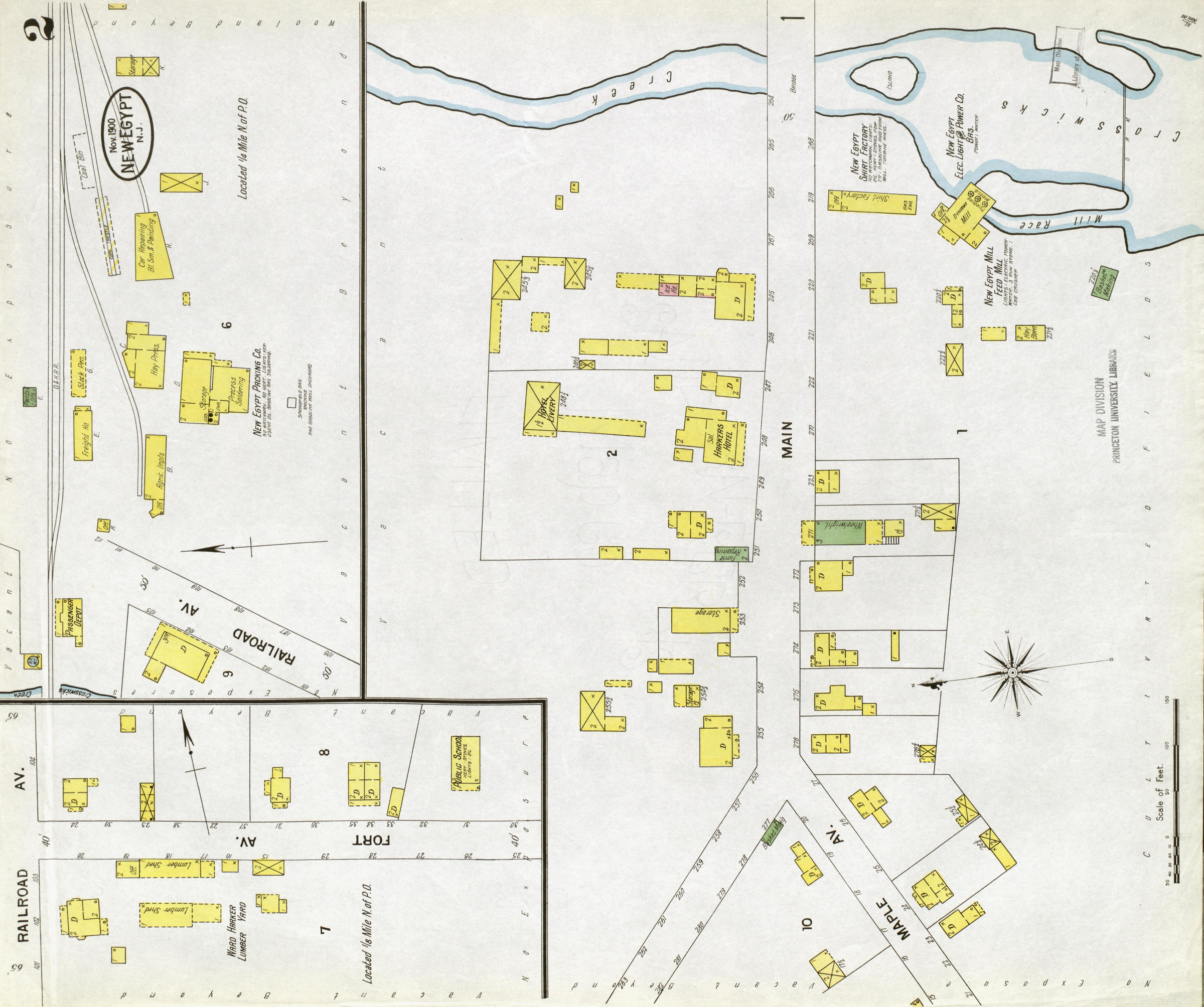}\\[.6em]
\includegraphics[width=\linewidth,height=.43\textheight,keepaspectratio]{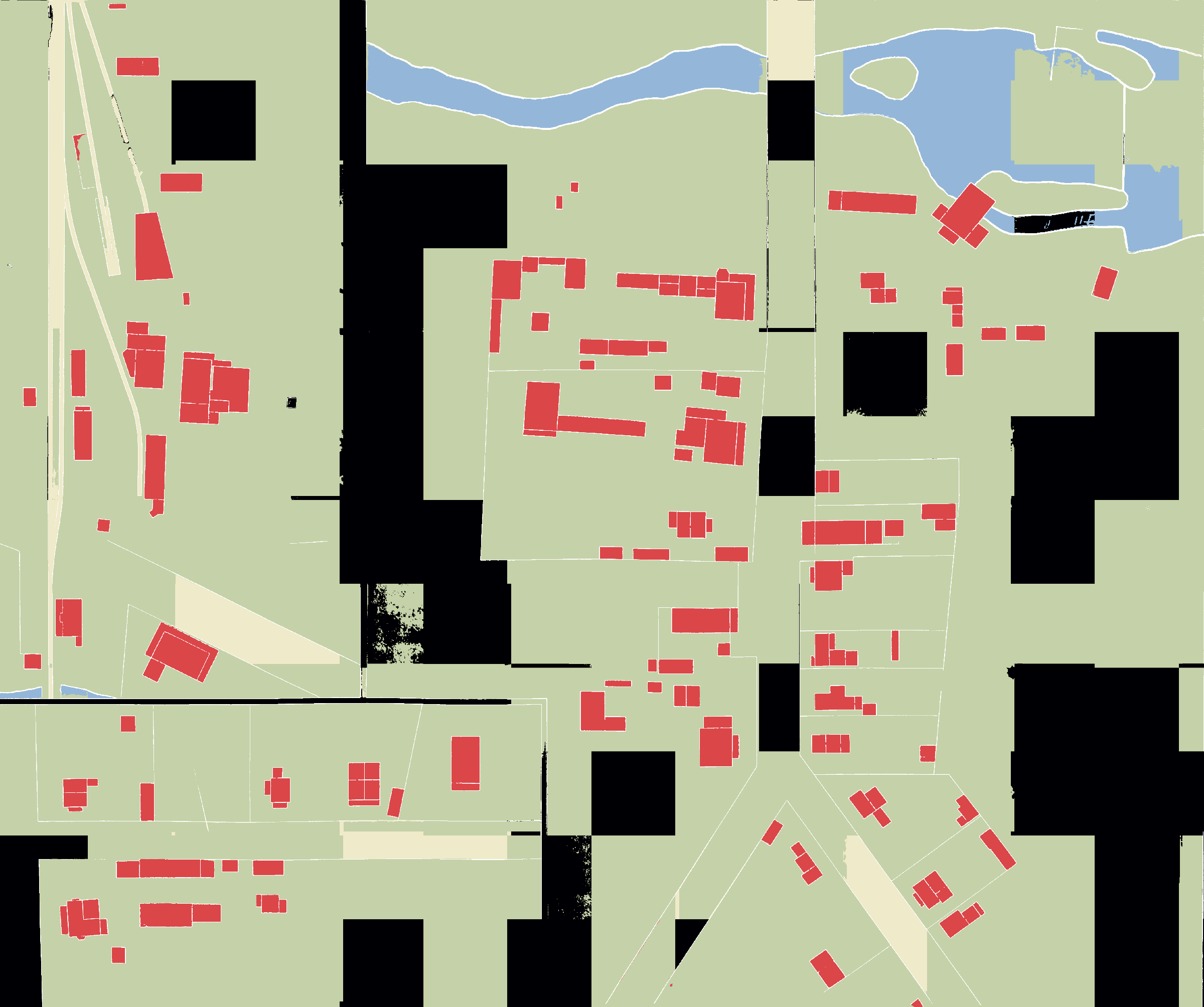}
\caption{(top) Fire insurance plan of New Egypt, Ocean, New Jersey, 1900. New Egypt N.J. 2, 1900. Sanborn Map Company, New York. 64 x 54 cm. Princeton library, HMC04. Rotated 90°. URL: \url{catalog.princeton.edu/catalog/9955740603506421}. (bottom) Result of the semantic segmentation. Building segmentation is accurate, whereas a lack of spatial context impairs the recognition of the road network and non-built surfaces.}
\label{fig:12}
\end{figure}

Figure~\ref{fig:7} presents inference details for a random subset of eight test samples. Figures~\ref{fig:8} through \ref{fig:12} portray full-image semantic segmentations for a selected set of five difficult cases. Four more typical examples are provided in Figures~\ref{fig:a1}–\ref{fig:a4}, in the Appendix. The qualitative results show that the detection of the boundary class remains unsatisfactory. This lower recognition performance also seems to marginally impact the recognition of other linear features, such as linear roads and waterways. By contrast, surfaces are generally well recognized.

One notable exception is Figure~\ref{fig:7}g. In that case, no distinctive cues help classify the empty surface as either non-built land or water/sea. When morphological or graphical information is available, as in Figures~\ref{fig:7}a and \ref{fig:8}, land can usually be distinguished from seas. The model appears to rely on color for the detection of water when the map is colored; this reliance explains, for instance, the inversion of water and landmasses in Figure~\ref{fig:9}, an ethnographic map from the 19\textsuperscript{th} century in which some land regions are tinted blue, leading to an inversion of sea and land in the segmentation mask. By contrast, iconographic content and allegories, like the marine creatures and ships depicted in Figure~\ref{fig:10}, are not always used effectively by the network. One counterexample is the view of Thomaston shown in Figure~\ref{fig:11}, where the map is well segmented despite its pictoriality.

Distinguishing linear features from one another can be challenging. One example is Figure~\ref{fig:7}b, where roads and rivers tend to be confused. In this case, the distinction relies mainly on context and on the graphical coherence between the sea and the river, both represented by horizontal hatchings. In Figure~\ref{fig:7}c, the Cogon River, drawn with a hatched blue line, is confused with a boundary line, a mistake that occurs quite frequently according to the confusion matrix (Fig.~\ref{fig:5}). In Figure~\ref{fig:7}f, however, the distinction between roads and waterways is generally accurate, probably owing to clear color differentiation and—possibly—hillshading, which might help the model recognize the rivers carving the landscape. In Figure~\ref{fig:7}d, the smaller paths are delimited only by a dotted line traversing a field that is itself depicted with a dotted texture. Although the model detects these paths, it does not classify them correctly, assigning them instead to the boundary class. Heavy texturing also impairs the recognition of other thin or small features, as in Figure~\ref{fig:7}e, where hachures significantly complicate the detection of dotted paths, rivers, and even houses. Features below a minimal area of a few pixels are generally hard to recognize, as observed with the cities in Figure~\ref{fig:7}f and the houses in Figure~\ref{fig:7}h. In the latter case, low contrast and high graphical density might also be partly responsible.

Figure~\ref{fig:8} constitutes an instructive example. In this cartogrammatic map of railway usage, statistical data are overlaid on a map of Europe, yet the model manages to overlook these strong graphical cues in several locations, notably along the German–Russian border, in French Normandy, and in the northwest of Ireland and the northeast of England. Figure~\ref{fig:12}, a Sanborn fire insurance map, illustrates a converse situation in which graphical density is too low; although the image resolution is high, graphical cues remain sparse, making roads and non-built areas hard to recognize.

\FloatBarrier

\section{General discussion of segmentation results}\label{sec:6}

The proposed method delivers strong performance across diverse map contexts and outperforms the reproduced baselines on both HCMSSD–Paris and HCMSSD–World. One important limitation remains the detection of boundaries and thin linear features, which was not the focus of this area-segmentation study. The boundary output is a semantic mask and converting generic historical-map boundaries into topologically valid, closed, instance-level geometries is a non-trivial scientific problem, which requires dedicated losses, post-processing, and evaluation. On coherent map series, dedicated architectures and strategies can be employed for precise deep edge filtering \cite{FP55GPVF}. The training of a separate, specialist model would therefore be necessary if the use case were to extract closed shapes, i.e., to vectorize the geometries. Overall, the Swin-based model segments map areas by leveraging visual context and multiscale integration at inference time. Where graphical cues are sparse, as in Figures~\ref{fig:10} and \ref{fig:12}, further integration scales could be useful; only two scales are implemented here. Beyond thin linear features, small objects such as icons may also be missed by the present architecture.

This research differs from the usual series-specific setting of map recognition by testing a generalizable model on heterogeneous training and benchmark data. In recent years, several studies have questioned the pertinence of specialist models for map recognition \cite{5A8MM6XB,CGK4FENK}. The controlled reproductions clarify the role of diversity and demonstrate the viability of generic pretraining: UNet–ResNet101, TransUNet, and HRNet–OCRNet each perform 7.6–12.5 mIoU points worse on HCMSSD–World than on the more homogeneous HCMSSD–Paris dataset, whereas Mask2Former–Swin-L maintains similar performance on both. Direct application of the generic checkpoint already reaches 74.4\% mIoU on World, and Semap-pretrained transfer reaches 76.3\%, compared with 74.2\% for target-specific retraining without Semap pretraining. On Paris, target-specific optimization raises performance from 71.5\% to 76.0\%, showing that coherent geographic context can still be learned profitably. Together, these results indicate that heterogeneous pretraining is especially useful for generic recognition, while dataset-specific optimization remains appropriate when a coherent target series is available.

Procedural generation was selected as an implementation choice because it provides explicit control over the frequency of textures and visual features and permits additional layers such as relief and place names. This control supports diverse training samples but does not guarantee equivalence with real historical maps: the result depends on expert-informed parameterization, and Figure~\ref{fig:4} identifies residual composition and visual-style differences that future generator versions should address.

\FloatBarrier

\section{Conclusion}\label{sec:7}

This article introduced Semap, an openly available dataset and benchmark consisting of 1,439 manually annotated map samples based on the ADHOC image database. It also evaluated a generic semantic-segmentation workflow, procedural synthetic pretraining, and multiscale integration. Under the harmonized benchmarks used here, the workflow achieves the strongest performance among the evaluated approaches. Synthetic and manually annotated training data, as well as the model checkpoint, are published in open access along with the article \cite{D9ZE29XW}.

The results support the viability of semantic segmentation across heterogeneous collections, production contexts, and cartographic styles. This stands in contrast with the dominant paradigm of specialist, series-specific models and indicates that diverse training data can support robust learning when paired with suitable architectures. The present research also shows that controlled procedural data generation can complement scarce annotations and expose the model to varied morphology and graphic conventions. The statistical comparison in Figure~\ref{fig:4} nevertheless identifies remaining composition and visual-style gaps, especially for water prevalence, contrast, edge density, and texture regularity. Procedural synthesis should therefore be treated as a controlled approximation of the real domain, while retaining explicit control over stylistic variation.

As illustrated by the qualitative examples and the full-map segmentation results in the Appendix, the model can already reliably segment a variety of maps. This creates immediate possibilities for historical geography and urban-economic research, provided that the resulting masks are georeferenced and critically validated: the ontology may support researchers interested in measuring historical hydrographic extent, modeling the relationship between road networks and urban growth, studying the expansion and morphology of built-up areas, the distribution of open or non-built land, or the spatial association of settlements with transport corridors and waterways. At the same time, the model can inform research on maps as historical artefacts. At corpus scale, the relative extent and arrangement of built, non-built, water, road, boundary, and background classes can support computational analyses of composition and framing, and address research questions in digital critical cartography concerning historical biases towards certain semantic-geographic subjects. Geographic-semantic profiles could also assist the thematic classification of material within large cartographic archives.

The present semantic boundary masks should not be interpreted as generic vector outputs: robust conversion to topologically valid instances remains an open research problem requiring dedicated methods and evaluation. With improved recognition of boundaries, thin features, and small objects, and with integration of georeferencing, text recognition, and temporal metadata, the broader approach could nevertheless support historical geographic information systems by making the long tail of cartographic archives more accessible as linked and queryable evidence. Such an environment would allow researchers to move across places and dates, compare alternative representations, inspect source maps behind extracted information, and trace changes in hydrography, land use, settlement, and transport networks.

\section*{Acknowledgements}

I wish to acknowledge the valuable contributions of Damien Donoso-Gomez, Ben Kriesel, and Danyang Wang, to the annotation of a portion of the Semap dataset. I further extend my thanks to Isabella di Lenardo, for the resources and support she offered to pursue this project.

\section*{Note on the use of AI in the manuscript preparation process}

Artificial intelligence (AI) agent technologies and language models were occasionally employed in the preparation of this manuscript, in particular DeepL’s translation models and OpenAI’s GPT-4o. These models were utilized for translation, text correction, reformulation, and proofreading. Conversely, AI was not employed for the production of scientific content, such as the description of the methodology, the analysis or the discussion of results, nor the formulation of conclusions.

\section*{Declaration of competing interests}

The author declares that he has no conflicts of interest.

\section*{Funding details}

This work was entirely funded by the public endowment of EPFL, Swiss Federal Institute of Technology in Lausanne.

\section*{Data availability statement}

The Semap release on Zenodo comprises 1,439 manually annotated map samples, 12,122 synthetically generated samples, the train/validation/test splits, and the resulting trained segmentation model \cite{D9ZE29XW}. The implementation and model configurations are available through the MMSegmentation repository cited in Section~\ref{sec:4.1}.

\FloatBarrier

\bibliography{references}

@article{2ED5JES3,
  author = {di Lenardo, Isabella and Barman, Rapha{\"{e}}l and Pardini, Federica and Kaplan, Fr{\'{e}}d{\'{e}}ric},
  title = {{Une approche computationnelle du cadastre napoléonien de Venise}},
  year = {2021},
  volume = {3},
  doi = {0.4000/revuehn.1786},
  journal = {Humanités numériques},
}

@misc{2YTNGM73,
  title = {{MapTiler Planet}},
  year = {n.d.},
  url = {https://www.maptiler.com/planet/},
  howpublished = {MapTiler},
  note = {Accessed 2025-4-19},
}

@inproceedings{3HCXCCDR,
  author = {Petitpierre, Remi},
  title = {{Mapping Memes in the Napoleonic Cadastre: Expanding Frontiers in Memetics}},
  year = {2023},
  pages = {3},
  doi = {10.5281/zenodo.8107916},
  publisher = {Zenodo},
  booktitle = {Digital Humanities 2023: Book of Abstracts},
}

@misc{4F8ZKYWC,
  title = {{MMSegmentation: OpenMMLab Semantic Segmentation Toolbox and Benchmark}},
  year = {2022},
  url = {https://github.com/open-mmlab/mmsegmentation},
  publisher = {OpenMMLab},
  note = {Version 0.28.0},
  howpublished = {Computer software},
}

@inproceedings{4H4XGM2Y,
  author = {Petitpierre, Remi and di Lenardo, Isabella and Rappo, Lucas},
  title = {{Revealing the Structure of Land Ownership through the Automatic Vectorisation of Swiss Cadastral Plans}},
  year = {2024},
  doi = {10.13140/RG.2.2.26632.33281},
  url = {https://digihistch24.github.io/submissions/454/},
  booktitle = {Digital History Switzerland},
}

@misc{4YXKU2KM,
  author = {Petitpierre, Remi},
  title = {{Historical City Maps Semantic Segmentation Dataset}},
  year = {2021},
  url = {https://doi.org/10.5281/zenodo.5497934},
  publisher = {Zenodo},
}

@misc{5A8MM6XB,
  author = {Kim, Jina and Li, Zekun and Lin, Yijun and Namgung, Min and Jang, Leeje and Chiang, Yao-Yi},
  title = {{The mapKurator System: A Complete Pipeline for Extracting and Linking Text from Historical Maps}},
  year = {2023},
  doi = {10.48550/arXiv.2306.17059},
  url = {http://arxiv.org/abs/2306.17059},
  publisher = {arXiv},
  note = {Accessed 2024-1-11},
}

@article{5BC2XM8P,
  author = {Uhl, Johannes H. and Leyk, Stefan and Chiang, Yao-Yi and Duan, Weiwei and Knoblock, Craig A.},
  title = {{Automated Extraction of Human Settlement Patterns From Historical Topographic Map Series Using Weakly Supervised Convolutional Neural Networks}},
  year = {2020},
  volume = {8},
  pages = {6978-6996},
  doi = {10.1109/access.2019.2963213},
  journal = {IEEE Access},
}

@inproceedings{63LS594U,
  author = {Oliveira, Sofia A. and di Lenardo, Isabella and Tourenc, Bastien and Kaplan, Frederic},
  title = {{A deep learning approach to Cadastral Computing}},
  year = {2019},
  url = {https://dev.clariah.nl/files/dh2019/boa/0691.html},
  booktitle = {Digital Humanities Conference},
  note = {Accessed 2020-5-22},
}

@article{6IKIYPW2,
  author = {Arzoumanidis, Lukas and Fethers, James Ormond and Mudiyanselage, Sethmiya Herath and Dehbi, Youness},
  title = {{Deep Generation of Synthetic Training Data for the Automated Extraction of Semantic Knowledge from Historical Maps}},
  year = {2024},
  volume = {7},
  pages = {1-2},
  doi = {10.5194/ica-abs-7-7-2024},
  journal = {Abstracts of the ICA},
}

@misc{7U3E4BRJ,
  author = {Dosovitskiy, Alexey and Beyer, Lucas and Kolesnikov, Alexander and Weissenborn, Dirk and Zhai, Xiaohua and Unterthiner, Thomas and Dehghani, Mostafa and Minderer, Matthias and Heigold, Georg and Gelly, Sylvain and Uszkoreit, Jakob and Houlsby, Neil},
  title = {{An Image is Worth 16x16 Words: Transformers for Image Recognition at Scale}},
  year = {2021},
  doi = {10.48550/arXiv.2010.11929},
  url = {http://arxiv.org/abs/2010.11929},
  publisher = {arXiv},
  note = {Accessed 2023-2-6},
}

@article{7XVY2CL5,
  author = {M{\"{a}}yr{\"{a}}, Janne and Kivinen, Sonja and Keski-Saari, Sarita and Poikolainen, Laura and Kumpula, Timo},
  title = {{Utilizing historical maps in identification of long-term land use and land cover changes}},
  year = {2023},
  doi = {10.1007/s13280-023-01838-z},
  url = {https://doi.org/10.1007/s13280-023-01838-z},
  journal = {Ambio},
  note = {Accessed 2023-3-15},
}

@misc{8Q9EIGGG,
  author = {Xia, Xue and Zhang, Daiwei and Song, Wenxuan and Huang, Wei and Hurni, Lorenz},
  title = {{MapSAM: Adapting Segment Anything Model for Automated Feature Detection in Historical Maps}},
  year = {2024},
  doi = {10.48550/arXiv.2411.06971},
  url = {http://arxiv.org/abs/2411.06971},
  publisher = {arXiv},
  note = {Accessed 2025-3-19},
}

@misc{9BVZMKVZ,
  author = {M{\"{u}}hlematter, Dominik J. and Schweizer, Sebastian and Jiao, Chenjing and Xia, Xue and Heitzler, Magnus and Hurni, Lorenz},
  title = {{Probabilistic road classification in historical maps using synthetic data and deep learning}},
  year = {2024},
  doi = {10.48550/arXiv.2410.02250},
  url = {http://arxiv.org/abs/2410.02250},
  publisher = {arXiv},
  note = {Accessed 2025-3-19},
}

@inproceedings{AGXDA7XV,
  author = {Liu, Ze and Lin, Yutong and Cao, Yue and Hu, Han and Wei, Yixuan and Zhang, Zheng and Lin, Stephen and Guo, Baining},
  title = {{Swin Transformer: Hierarchical Vision Transformer using Shifted Windows}},
  year = {2021},
  pages = {9992-10002},
  doi = {10.1109/ICCV48922.2021.00986},
  url = {https://www.computer.org/csdl/proceedings-article/iccv/2021/281200j992/1BmGKZoEzug},
  publisher = {IEEE Computer Society},
  booktitle = {2021 IEEE/CVF International Conference on Computer Vision (ICCV)},
  note = {Accessed 2025-4-19},
}

@article{AWGGZTC7,
  author = {Maxwell, Aaron E. and Bester, Michelle S. and Guillen, Luis A. and Ramezan, Christopher A. and Carpinello, Dennis J. and Fan, Yiting and Hartley, Faith M. and Maynard, Shannon M. and Pyron, Jaimee L.},
  title = {{Semantic Segmentation Deep Learning for Extracting Surface Mine Extents from Historic Topographic Maps}},
  year = {2020},
  volume = {12},
  number = {24},
  pages = {4145},
  doi = {10.3390/rs12244145},
  journal = {Remote Sensing},
}

@article{BGEQN6UN,
  author = {Grabska-Szwagrzyk, Ewa and Jakiel, Michał and Keeton, William and Kozak, Jacek and Kuemmerle, Tobias and Onoszko, Kamil and Ostafin, Krzysztof and Shahbandeh, Mahsa and Szubert, Piotr and Szwagierczak, Anna and Szwagrzyk, Jerzy and Zi{\'{o}}łkowska, El{\.{z}}bieta and Kaim, Dominik},
  title = {{Historical maps improve the identification of forests with potentially high conservation value}},
  year = {2024},
  volume = {17},
  number = {5},
  pages = {e13043},
  doi = {10.1111/conl.13043},
  journal = {Conservation Letters},
}

@article{BMBEHXRD,
  author = {Liu, Guilin and Li, Jingyun and Nie, Peng},
  title = {{Tracking the history of urban expansion in Guangzhou (China) during 1665–2017: Evidence from historical maps and remote sensing images}},
  year = {2022},
  volume = {112},
  pages = {105773},
  doi = {10.1016/j.landusepol.2021.105773},
  journal = {Land Use Policy},
}

@article{BSB33Q7P,
  author = {Xia, Xue and Zhang, Tao and Heitzler, Magnus and Hurni, Lorenz},
  title = {{Vectorizing historical maps with topological consistency: A hybrid approach using transformers and contour-based instance segmentation}},
  year = {2024},
  volume = {129},
  pages = {103837},
  doi = {10.1016/j.jag.2024.103837},
  journal = {International Journal of Applied Earth Observation and Geoinformation},
}

@article{CBST88UR,
  author = {Jiao, Chenjing and Heitzler, Magnus and Hurni, Lorenz},
  title = {{A survey of road feature extraction methods from raster maps}},
  year = {2021},
  volume = {25},
  number = {6},
  pages = {2734-2763},
  doi = {10.1111/tgis.12812},
  journal = {Transactions in GIS},
}

@mastersthesis{CGK4FENK,
  author = {Petitpierre, Remi},
  title = {{Neural networks for semantic segmentation of historical city maps: Cross-cultural performance and the impact of figurative diversity}},
  year = {2020},
  publisher = {EPFL},
  school = {EPFL},
}

@article{D33PL94I,
  author = {Arzoumanidis, Lukas and Knechtel, Julius and Haunert, Jan-Henrik and Dehbi, Youness},
  title = {{Semantic segmentation of historical maps using Self-Constructing Graph Convolutional Networks}},
  year = {2025},
  pages = {1-11},
  doi = {10.1080/15230406.2025.2468304},
  journal = {Cartography and Geographic Information Science},
}

@article{DMHFPAJU,
  author = {Petitpierre, Remi and Guhennec, Paul},
  title = {{Effective annotation for the automatic vectorization of cadastral maps}},
  year = {2023},
  volume = {38},
  number = {3},
  pages = {1227-1237},
  doi = {10.1093/llc/fqad006},
  journal = {Digital Scholarship in the Humanities},
}

@inproceedings{ESMCSY4S,
  author = {Ronneberger, Olaf and Fischer, Philipp and Brox, Thomas},
  title = {{U-net: Convolutional networks for biomedical image segmentation}},
  year = {2015},
  volume = {9351},
  pages = {234--241},
  doi = {10.1007/978-3-319-24574-4_28},
  publisher = {Springer},
  booktitle = {Lecture Notes in Computer Science},
}

@article{ETSAEFRC,
  author = {Martinez, Th{\'{e}}o and Hammoumi, Adam and Ducret, Gabriel and Moreaud, Maxime and Deschamps, R{\'{e}}my and Herv{\'{e}}, Piegay and Berger, Jean-Fran{\c{c}}ois},
  title = {{Deep learning ancient map segmentation to assess historical landscape changes}},
  year = {2023},
  volume = {19},
  number = {1},
  pages = {2225071},
  doi = {10.1080/17445647.2023.2225071},
  journal = {Journal of Maps},
}

@inproceedings{FP55GPVF,
  author = {Chen, Yizi and Carlinet, Edwin and Chazalon, Joseph and Mallet, Cl{\'{e}}ment and Dum{\'{e}}nieu, Bertrand and Perret, Julien},
  title = {{Vectorization of Historical Maps Using Deep Edge Filtering and Closed Shape Extraction}},
  year = {2021},
  pages = {510--525},
  publisher = {Springer},
  booktitle = {Document Analysis and Recognition – ICDAR 2021},
}

@misc{GYRSWYW7,
  title = {{Mapbox Terrain v2 | Tilesets | Mapbox Docs}},
  year = {n.d.},
  url = {https://docs.mapbox.com//data/tilesets/reference/mapbox-terrain-v2},
  howpublished = {Mapbox},
  note = {Accessed 2025-4-19},
}

@techreport{HXWMWXM9,
  author = {Li, Shanci and Cerioni, Alessandro and Herny, Cl{\'{e}}mence and Duriaux, Henrich and Pott, Roxane},
  title = {{Vectorization of historical cadastral plans from the 1850s in the Canton of Geneva}},
  year = {2024},
  url = {https://tech.stdl.ch/PROJ-CADMAP/},
  publisher = {Swiss Territorial Data Lab},
  institution = {Swiss Territorial Data Lab},
  note = {Accessed 2025-4-19},
}

@inproceedings{IBI9YNCQ,
  author = {Yuan, Yuhui and Chen, Xilin and Wang, Jingdong},
  title = {{Object-Contextual Representations for Semantic Segmentation}},
  year = {2020},
  pages = {173-190},
  doi = {10.1007/978-3-030-58539-6_11},
  publisher = {Springer},
  booktitle = {Computer Vision – ECCV 2020},
}

@incollection{IKTPDJBV,
  author = {McDonough, Katherine},
  title = {{Maps as Data}},
  year = {2024},
  publisher = {University of Minnesota Press},
  booktitle = {Computational Humanities:  Debates in Digital Humanities},
}

@article{JHEVKQ8S,
  author = {Ekim, Burak and Sertel, Elif and Kabadayı, M. Erdem},
  title = {{Automatic Road Extraction from Historical Maps Using Deep Learning Techniques: A Regional Case Study of Turkey in a German World War II Map}},
  year = {2021},
  volume = {10},
  number = {8},
  pages = {492},
  doi = {10.3390/ijgi10080492},
  journal = {ISPRS International Journal of Geo-Information},
}

@article{JZMFGI49,
  author = {Uhl, Johannes H. and Leyk, Stefan and Chiang, Yao-Yi and Knoblock, Craig A.},
  title = {{Towards the automated large-scale reconstruction of past road networks from historical maps}},
  year = {2022},
  volume = {94},
  pages = {101794},
  doi = {10.1016/j.compenvurbsys.2022.101794},
  journal = {Computers, Environment and Urban Systems},
}

@article{KPIDQQ4F,
  author = {Chen, Yizi and Chazalon, Joseph and Carlinet, Edwin and Ngoc, Minh {\^{O}}n V{\~{u}} and Mallet, Cl{\'{e}}ment and Perret, Julien},
  title = {{Automatic vectorization of historical maps: A benchmark}},
  year = {2024},
  volume = {19},
  number = {2},
  pages = {e0298217},
  doi = {10.1371/journal.pone.0298217},
  journal = {PLOS ONE},
}

@misc{LECIPW5W,
  author = {Kirillov, Alexander and Mintun, Eric and Ravi, Nikhila and Mao, Hanzi and Rolland, Chloe and Gustafson, Laura and Xiao, Tete and Whitehead, Spencer and Berg, Alexander C. and Lo, Wan-Yen and Doll{\'{a}}r, Piotr and Girshick, Ross},
  title = {{Segment Anything}},
  year = {2023},
  doi = {10.48550/arXiv.2304.02643},
  url = {http://arxiv.org/abs/2304.02643},
  publisher = {arXiv},
  note = {Accessed 2024-1-24},
}

@article{LVXMMF5T,
  author = {Heitzler, Magnus and Hurni, Lorenz},
  title = {{Cartographic reconstruction of building footprints from historical maps: A study on the Swiss Siegfried map}},
  year = {2020},
  volume = {24},
  number = {2},
  pages = {442--461},
  doi = {10.1111/tgis.12610},
  journal = {Transactions in GIS},
}

@article{LZ5GNULF,
  author = {Jiao, Chenjing and Heitzler, Magnus and Hurni, Lorenz},
  title = {{A fast and effective deep learning approach for road extraction from historical maps by automatically generating training data with symbol reconstruction}},
  year = {2022},
  volume = {113},
  pages = {102980},
  doi = {10.1016/j.jag.2022.102980},
  journal = {International Journal of Applied Earth Observation and Geoinformation},
}

@misc{MN8LTGQS,
  author = {Chazalon, Joseph and Carlinet, Edwin and Chen, Yizi and Perret, Julien and Dum{\'{e}}nieu, Bertrand and Mallet, Cl{\'{e}}ment and G{\'{e}}raud, Thierry and Nguyen, Vincent and Nguyen, Nam and Baloun, Josef and Lenc, Ladislav and Kr{\'{a}}l, Pavel},
  title = {{ICDAR 2021 Competition on Historical Map Segmentation}},
  year = {2021},
  doi = {10.48550/arXiv.2105.13265},
  publisher = {arXiv},
  note = {Accessed 2021-7-1},
}

@article{MUL2WJ9M,
  author = {Zhao, Yao and Wang, Guangxia and Yang, Jian and Li, Tingting and Li, Ziwei},
  title = {{AU3-GAN: A Method for Extracting Roads from Historical Maps Based on an Attention Generative Adversarial Network}},
  year = {2024},
  volume = {8},
  number = {2},
  pages = {26},
  doi = {10.1007/s41651-024-00187-z},
  journal = {Journal of Geovisualization and Spatial Analysis},
}

@inproceedings{P32JKYH2,
  author = {Uhl, Johannes H. and Leyk, Stefan and Chiang, Yao-Yi and Duan, Weiwei and Knoblock, Craig A.},
  title = {{Extracting human settlement footprint from historical topographic map series using context-based machine learning}},
  year = {2017},
  pages = {1-6},
  doi = {10.1049/cp.2017.0144},
  booktitle = {8th ​​International Conference of Pattern Recognition Systems (ICPRS 2017)},
}

@misc{P4FDQUVK,
  author = {Yuan, Yunshuang and Thiemann, Frank and Sester, Monika},
  title = {{Semantic Segmentation for Sequential Historical Maps by Learning from Only One Map}},
  year = {2025},
  doi = {10.48550/arXiv.2501.01845},
  url = {http://arxiv.org/abs/2501.01845},
  publisher = {arXiv},
  note = {Accessed 2025-4-19},
}

@phdthesis{PJC6ZP3W,
  author = {Petitpierre, Remi},
  title = {{Studying Maps at Scale: A Digital Investigation of Cartography and the Evolution of Figuration}},
  year = {2025},
  publisher = {EPFL},
  school = {EPFL},
}

@inproceedings{PKWUZHSU,
  author = {Wu, Sidi and Chen, Yizi and Schindler, Konrad and Hurni, Lorenz},
  title = {{Cross-attention Spatio-temporal Context Transformer for Semantic Segmentation of Historical Maps}},
  year = {2023},
  pages = {1--9},
  doi = {10.1145/3589132.3625572},
  url = {https://doi.org/10.1145/3589132.3625572},
  publisher = {Association for Computing Machinery},
  booktitle = {Proceedings of the 31st ACM International Conference on Advances in Geographic Information Systems},
  note = {Accessed 2025-3-19},
}

@article{PU5SZPTU,
  author = {Levin, Gregor and Groom, Geoff and Svenningsen, Stig Roar},
  title = {{Assessing spatially explicit long-term landscape dynamics based on automated production of land category layers from Danish late nineteenth-century topographic maps in comparison with contemporary maps}},
  year = {2025},
  volume = {197},
  number = {2},
  pages = {195},
  doi = {10.1007/s10661-025-13634-1},
  journal = {Environmental Monitoring and Assessment},
}

@article{PVXGKXCR,
  author = {Uhl, Johannes H. and Leyk, Stefan and Li, Zekun and Duan, Weiwei and Shbita, Basel and Chiang, Yao-Yi and Knoblock, Craig A.},
  title = {{Combining Remote-Sensing-Derived Data and Historical Maps for Long-Term Back-Casting of Urban Extents}},
  year = {2021},
  volume = {13},
  number = {18},
  pages = {3672},
  doi = {10.3390/rs13183672},
  journal = {Remote Sensing},
}

@article{QDTIFG72,
  author = {Can, Yekta Said and Gerrits, Petrus Johannes and Kabadayi, M. Erdem},
  title = {{Automatic Detection of Road Types From the Third Military Mapping Survey of Austria-Hungary Historical Map Series With Deep Convolutional Neural Networks}},
  year = {2021},
  volume = {9},
  pages = {62847-62856},
  doi = {10.1109/ACCESS.2021.3074897},
  journal = {IEEE Access},
}

@mastersthesis{QWHQBJDN,
  author = {Jan, Maxime F.},
  title = {{A generic method for cartographic realignment using local feature matching: towards a computational urban history}},
  year = {2022},
  publisher = {EPFL},
  school = {EPFL},
}

@inproceedings{R6RP2WMJ,
  author = {Hosseini, Kasra and Wilson, Daniel C. S. and Beelen, Kaspar and McDonough, Katherine},
  title = {{MapReader: a computer vision pipeline for the semantic exploration of maps at scale}},
  year = {2022},
  pages = {8--19},
  doi = {10.1145/3557919.3565812},
  url = {https://doi.org/10.1145/3557919.3565812},
  publisher = {Association for Computing Machinery},
  booktitle = {Proceedings of the 6th ACM SIGSPATIAL International Workshop on Geospatial Humanities},
  note = {Accessed 2022-11-22},
}

@article{R943LAEU,
  author = {Arzoumanidis, Lukas and Knechtel, Julius and Haunert, Jan-Henrik and Dehbi, Youness},
  title = {{Self-Constructing Graph Convolutional Networks for Semantic Segmentation of Historical Maps}},
  year = {2023},
  volume = {6},
  pages = {1-2},
  doi = {10.5194/ica-abs-6-11-2023},
  journal = {Abstracts of the ICA},
}

@misc{RZI93QMS,
  author = {Li, Zekun and Guan, Runyu and Yu, Qianmu and Chiang, Yao-Yi and Knoblock, Craig A.},
  title = {{Synthetic Map Generation to Provide Unlimited Training Data for Historical Map Text Detection}},
  year = {2021},
  doi = {10.1145/3486635.3491070},
}

@inproceedings{S2XGZJE4,
  author = {Li, Zekun},
  title = {{Generating Historical Maps from Online Maps}},
  year = {2019},
  pages = {610--611},
  doi = {10.1145/3347146.3363463},
  url = {https://dl.acm.org/doi/10.1145/3347146.3363463},
  publisher = {Association for Computing Machinery},
  booktitle = {Proceedings of the 27th ACM SIGSPATIAL International Conference on Advances in Geographic Information Systems},
  note = {Accessed 2025-4-18},
}

@inproceedings{S7WQIMQ4,
  author = {Xu, Yifan and Xu, Weijian and Cheung, David and Tu, Zhuowen},
  title = {{Line Segment Detection Using Transformers without Edges}},
  year = {2021},
  pages = {4255-4264},
  doi = {10.1109/CVPR46437.2021.00424},
  url = {https://www.computer.org/csdl/proceedings-article/cvpr/2021/4.509E260/1yeI8wB6bzW},
  publisher = {IEEE Computer Society},
  booktitle = {2021 IEEE/CVF Conference on Computer Vision and Pattern Recognition (CVPR)},
  note = {Accessed 2025-4-19},
}

@inproceedings{TI8VL2E4,
  author = {Jiao, Chenjing and Heitzler, Magnus and Hurni, Lorenz},
  title = {{A Novel Data Augmentation Method to Enhance the Training Dataset for Road Extraction from Swiss Historical Maps}},
  year = {2022},
  volume = {V-2-2022},
  pages = {423-429},
  doi = {10.5194/isprs-annals-V-2-2022-423-2022},
  url = {https://www.isprs-ann-photogramm-remote-sens-spatial-inf-sci.net/V-2-2022/423/2022/},
  publisher = {Copernicus GmbH},
  booktitle = {ISPRS Annals of the Photogrammetry, Remote Sensing and Spatial Information Sciences},
  note = {Accessed 2023-3-15},
}

@misc{TQADKIQH,
  author = {Cheng, Bowen and Misra, Ishan and Schwing, Alexander G. and Kirillov, Alexander and Girdhar, Rohit},
  title = {{Masked-attention Mask Transformer for Universal Image Segmentation}},
  year = {2022},
  doi = {10.48550/arXiv.2112.01527},
  url = {http://arxiv.org/abs/2112.01527},
  publisher = {arXiv},
  note = {Accessed 2023-11-30},
}

@article{TV7XH2BW,
  author = {G{\"{o}}derle, Wolfgang Thomas and Rampetsreiter, Fabian and Macher, Christian and Mauthner, Katrin and Pimas, Oliver},
  title = {{Deep learning for historical Cadastral maps and satellite imagery analysis: insights from Styria's Franciscean Cadastre}},
  year = {2024},
  volume = {18},
  number = {3},
  url = {https://dhq.digitalhumanities.org/vol/18/2/000744/000744.html},
  journal = {Digital humanities quarterly},
  note = {Accessed 2025-3-19},
}

@inproceedings{U23QD4TD,
  author = {Heitzler, Magnus and Hurni, Lorenz},
  title = {{Extracting Buildings from Historical Maps With Convolutional Neural Networks}},
  year = {2019},
  url = {https://www.research-collection.ethz.ch/handle/20.500.11850/371976},
  booktitle = {1st Swiss Workshop on Machine Learning for Environmental and Geosciences (MLEG 2019)},
  note = {Accessed 2024-7-24},
}

@misc{UYYWZK49,
  author = {He, Kaiming and Zhang, Xiangyu and Ren, Shaoqing and Sun, Jian},
  title = {{Deep Residual Learning for Image Recognition}},
  year = {2015},
  doi = {10.48550/arXiv.1512.03385},
  publisher = {arXiv},
  note = {Accessed 2019-11-12},
}

@article{UZ9LB7IU,
  author = {Wang, Jingdong and Sun, Ke and Cheng, Tianheng and Jiang, Borui and Deng, Chaorui and Zhao, Yang and Liu, Dong and Mu, Yadong and Tan, Mingkui and Wang, Xinggang and Liu, Wenyu and Xiao, Bin},
  title = {{Deep High-Resolution Representation Learning for Visual Recognition}},
  year = {2021},
  volume = {43},
  pages = {3349-3364},
  doi = {10.1109/TPAMI.2020.2983686},
  journal = {IEEE Transactions on Pattern Analysis and Machine Intelligence},
}

@article{VPHRFJP5,
  author = {Vynikal, Jakub and M{\"{u}}llerov{\'{a}}, Jana and Pacina, Jan},
  title = {{Deep learning approaches for delineating wetlands on historical topographic maps}},
  year = {2024},
  volume = {28},
  number = {5},
  pages = {1400-1411},
  doi = {10.1111/tgis.13193},
  journal = {Transactions in GIS},
}

@article{VU7QJ78P,
  author = {Christophe, Sidonie and Mermet, Samuel and Laurent, Morgan and Touya, Guillaume},
  title = {{Neural map style transfer exploration with GANs}},
  year = {2022},
  volume = {8},
  number = {1},
  pages = {18-36},
  doi = {10.1080/23729333.2022.2031554},
  journal = {International Journal of Cartography},
}

@inproceedings{VYLMUW8M,
  author = {Xie, Enze and Wang, Wenhai and Yu, Zhiding and Anandkumar, Anima and Alvarez, Jose M. and Luo, Ping},
  title = {{SegFormer: Simple and Efficient Design for Semantic Segmentation with Transformers}},
  year = {2021},
  url = {https://openreview.net/forum?id=OG18MI5TRL},
  publisher = {Remote},
  booktitle = {Advances in Neural Information Processing Systems},
  note = {Accessed 2025-4-19},
}

@article{W9UTP692,
  author = {Kang, Yuhao and , Gao ,Song and and Roth, Robert E.},
  title = {{Transferring multiscale map styles using generative adversarial networks}},
  year = {2019},
  volume = {5},
  number = {2-3},
  pages = {115-141},
  doi = {10.1080/23729333.2019.1615729},
  journal = {International Journal of Cartography},
}

@misc{XCDIEM9C,
  author = {G{\"{o}}derle, Wolfgang T and Macher, Christian and Mauthner, Katrin and Pimas, Oliver and Rampetsreiter, Fabian},
  title = {{AI-driven Structure Detection and Information Extraction from Historical Cadastral Maps (Early 19th Century Franciscean Cadastre in the Province of Styria) and Current High-resolution Satellite and Aerial Imagery for Remote Sensing}},
  year = {2023},
  doi = {10.48550/arXiv.2312.07560},
  url = {http://arxiv.org/abs/2312.07560},
  publisher = {arXiv},
  note = {Accessed 2024-7-9},
}

@misc{Y84567UY,
  author = {Zou, Mengjie and Dai, Tianhao and Petitpierre, R{\'{e}}mi and Vaienti, Beatrice and Kaplan, Fr{\'{e}}d{\'{e}}ric and Lenardo, Isabella},
  title = {{Recognizing and Sequencing Multi-word Texts in Maps Using an Attentive Pointer}},
  year = {2025},
  doi = {10.21203/rs.3.rs-6330456/v1},
  publisher = {In Review},
  note = {Accessed 2025-4-16},
}

@book{ZVQFDMWJ,
  author = {Chiang, Yao-Yi and Duan, Weiwei and Leyk, Stefan and Uhl, Johannes H. and Knoblock, Craig A.},
  title = {{Using Historical Maps in Scientific Studies: Applications, Challenges, and Best Practices}},
  year = {2020},
  url = {http://link.springer.com/10.1007/978-3-319-66908-3},
  publisher = {Springer},
  note = {Accessed 2020-5-6},
}

@inproceedings{ZYYEWSAX,
  author = {Xia, Xue and Jiao, Chenjing and Hurni, Lorenz},
  title = {{Contrastive Pretraining for Railway Detection: Unveiling Historical Maps with Transformers}},
  year = {2023},
  pages = {30--33},
  doi = {10.1145/3615886.3627738},
  url = {https://dl.acm.org/doi/10.1145/3615886.3627738},
  publisher = {Association for Computing Machinery},
  booktitle = {Proceedings of the 6th ACM SIGSPATIAL International Workshop on AI for Geographic Knowledge Discovery},
  note = {Accessed 2025-3-19},
}

@inproceedings{JUK5Y27S,
  author = {Milletari, Fausto and Navab, Nassir and Ahmadi, Seyed-Ahmad},
  title = {{V-Net: Fully Convolutional Neural Networks for Volumetric Medical Image Segmentation}},
  booktitle = {2016 Fourth International Conference on 3D Vision (3DV)},
  year = {2016},
  pages = {565-571},
  doi = {10.1109/3DV.2016.79},
  url = {https://ieeexplore.ieee.org/document/7785132},
}

@inproceedings{IT57HISL,
  author = {Loshchilov, Ilya and Hutter, Frank},
  title = {{Decoupled Weight Decay Regularization}},
  booktitle = {International Conference on Learning Representations},
  year = {2018},
  url = {https://openreview.net/forum?id=Bkg6RiCqY7},
}

@misc{D9ZE29XW,
  author = {Petitpierre, Remi and Gomez Donoso, Damien and Kriesel, Ben},
  title = {{Semantic Segmentation Map Dataset (Semap)}},
  year = {2025},
  publisher = {EPFL},
  doi = {10.5281/zenodo.16164781},
}

@inproceedings{U4F2XWXA,
  author = {Petitpierre, Remi and Kaplan, Frédéric and di Lenardo, Isabella},
  title = {{Generic Semantic Segmentation of Historical Maps}},
  booktitle = {Proceedings of the Conference on Computational Humanities Research 2021},
  year = {2021},
  volume = {2989},
  publisher = {CEUR},
  location = {Amsterdam, The Netherlands},
  url = {http://ceur-ws.org/Vol-2989/long_paper27.pdf},
}

@misc{Polak,
  author = {Pol\'ak, Matej},
  title = {{Segmentation of historical maps by deep learning}},
  year = {2024},
  school = {Czech Technical University},
  url = {https://dspace.cvut.cz/bitstream/handle/10467/115578/F8-BP-2024-Polak-Matej-thesis.pdf}
}

\clearpage

\appendix

\setcounter{figure}{0}

\setcounter{table}{0}

\renewcommand{\thefigure}{A\arabic{figure}}

\renewcommand{\thetable}{A\arabic{table}}

\renewcommand{\theHfigure}{A.\arabic{figure}}

\renewcommand{\theHtable}{A.\arabic{table}}

\section{Supplementary Materials}

\begin{table}[pos=htbp]
\centering
\scriptsize
\begin{adjustbox}{max width=\linewidth}
\begin{tabular}{@{}p{.23\linewidth}p{.18\linewidth}p{.16\linewidth}p{.09\linewidth}p{.12\linewidth}r@{}}
\toprule
Institution(s) & Part of & Landscape & Scale & Period & \# samples \\
\midrule
Vaud Cantonal Archives & Lausanne cadastre & Urban, periurban & very high & 1722–1882 & 42 \\
Neuchatel Cadastre Office & Neuchatel cadastre & Urban, periurban & very high & 1886–1888 & 10 \\
Geneva Cadastre Office & Geneva cadastre & Urban, periurban & very high & 1845–1856 & 8 \\
French Dept. Archives \& others & Napoleonic cadastre & Urban, rural & very high & 1808–1888 & 18 \\
BHVP & Paris Atlas & Urban & very high & 1898–1926 & 3 \\
BHVP & HCMSSD – Paris & Urban & very high & 1785–1950 & 70 \\
BnF & HCMSSD – Paris & Urban & high & 1765–1949 & 189 \\
(32 institutions) & HCMSSD – World & Urban, periurban & high & 1701–1949 & 97 \\
Berkeley library & Semap & diverse & low–high & 1657–1946 & 8 \\
BnF & Semap & diverse & low–high & 1494–1947 & 262 \\
U Bordeaux-Montaigne & Semap & diverse & low–high & 1676–1946 & 11 \\
Boston Leventhal Map Centre & Semap & diverse & low–very high & 1493–1947 & 17 \\
David Rumsey & Semap & diverse & low–high & 1525–1947 & 355 \\
e-rara & Semap & diverse & low–high & 1513–1947 & 47 \\
Library of Congress & Semap & diverse & low–very high & 1507–1947 & 165 \\
NY Public Library & Semap & diverse & low–very high & 1542–1947 & 44 \\
Princeton Library & Semap & diverse & low–high & 1509–1947 & 28 \\
U Leiden & Semap & diverse & low–high & 1545–1947 & 40 \\
WWU Munster & Semap & diverse & low–high & 1550–1938 & 12 \\
(5 other institutions) & Semap & diverse & low–high & 1492–1947 & 13 \\
(50+ institutions) & Semap & diverse & low–very high & 1492–1950 & \textbf{1439} \\
\bottomrule
\end{tabular}

\end{adjustbox}
\caption{Composition of Semap dataset.}
\label{tab:a1}
\end{table}

\begin{table}[pos=htbp]
\centering
\footnotesize
\begin{adjustbox}{max width=\linewidth}
\begin{tabular}{@{}p{.29\linewidth}lp{.64\linewidth}@{}}
\toprule
Sample-normalized macro average &  & $\displaystyle mIoU=\frac{1}{K}\sum _{k=3}^{6} \frac{\sum _{i=1}^{N} {IoU}_{i,k}\ |{Y}_{i,k}|}{\sum _{i=1}^{N} |{Y}_{i,k}|}$ \\
Mean per-image micro average &  & $\displaystyle mIoU=\frac{1}{N}\sum _{i=1}^{N} \frac{\sum _{k=3}^{6} \sum _{\forall p} {\cap }_{i,k}{Y}_{i,k}}{\sum _{k=3}^{6} \sum _{\forall p} {{\cup }_{i,k}Y}_{i,k}}$ \\
Global micro average &  & $\displaystyle mIoU=\frac{\sum _{k=3}^{6} \sum _{i=1}^{N} \sum _{\forall p} {\cap }_{i,k}}{\sum _{k=3}^{6} \sum _{i=1}^{N} \sum _{\forall p} {\cup }_{i,k}}$ \\
Per class macro average &  & $\displaystyle mIoU=\frac{1}{K}\sum _{k=3}^{6} \frac{\sum _{i=1}^{N} \sum _{\forall p} {\cap }_{i,k}}{\sum _{i=1}^{N} \sum _{\forall p} {\cup }_{i,k}}$ \\
\bottomrule
\end{tabular}

\end{adjustbox}
\caption{Performance assessment strategies. Variant formulae for computing mIoU. For mR, the union term in the denominator is replaced by the ground-truth pixel count; for mP, it is replaced by the predicted-pixel count.}
\label{tab:a2}
\end{table}

\begin{table}[pos=htbp]
\centering
\footnotesize
\begin{adjustbox}{max width=\linewidth}
\begin{tabular}{@{}p{.12\linewidth}p{.40\linewidth}rrrr@{}}
\toprule
Set & Performance assessment strategy & mIoU & mR & mP & mPR \\
\midrule
Semap & Sample-normalized macro average (base) & 74.2 & 79.4 & \textbf{85.4} & \textbf{82.4} \\
 & Mean per-image micro average & \textbf{74.4} & \textbf{80.3} & 80.3 & 80.3 \\
 & Global micro average & 67.0 & \textbf{80.3} & 80.3 & 80.3 \\
 & Per class macro average & 57.4 & 79.4 & 67.8 & 73.6 \\
HCMSSD Paris & Sample-normalized macro average & 79.3 & 84.2 & 87.7 & 85.9 \\
 & Mean per-image micro average & \textbf{84.6} & \textbf{90.6} & \textbf{90.6} & \textbf{90.6} \\
 & Global micro average & 82.8 & \textbf{90.6} & \textbf{90.6} & \textbf{90.6} \\
 & Per class macro average (base) & 76.0 & 84.2 & 88.6 & 86.4 \\
HCMSSD World & Sample-normalized macro average & \textbf{81.4} & 86.8 & \textbf{92.0} & \textbf{89.4} \\
 & Mean per-image micro average & 80.7 & \textbf{87.8} & 87.8 & 87.8 \\
 & Global micro average & 78.2 & \textbf{87.8} & 87.8 & 87.8 \\
 & Per class macro average (base) & 74.2 & 86.8 & 84.1 & 85.4 \\
\bottomrule
\end{tabular}

\end{adjustbox}
\caption{Comparison of performance assessment strategies on Semap, HCMSSD–Paris, and –World datasets (retrained models). Based on the formulae provided in Tab.~\ref{tab:a2}. mPR = (mP + mR)/2.}
\label{tab:a3}
\end{table}

\begin{figure}[pos=htbp,align=\centering]
\includegraphics[width=\linewidth,height=.43\textheight,keepaspectratio]{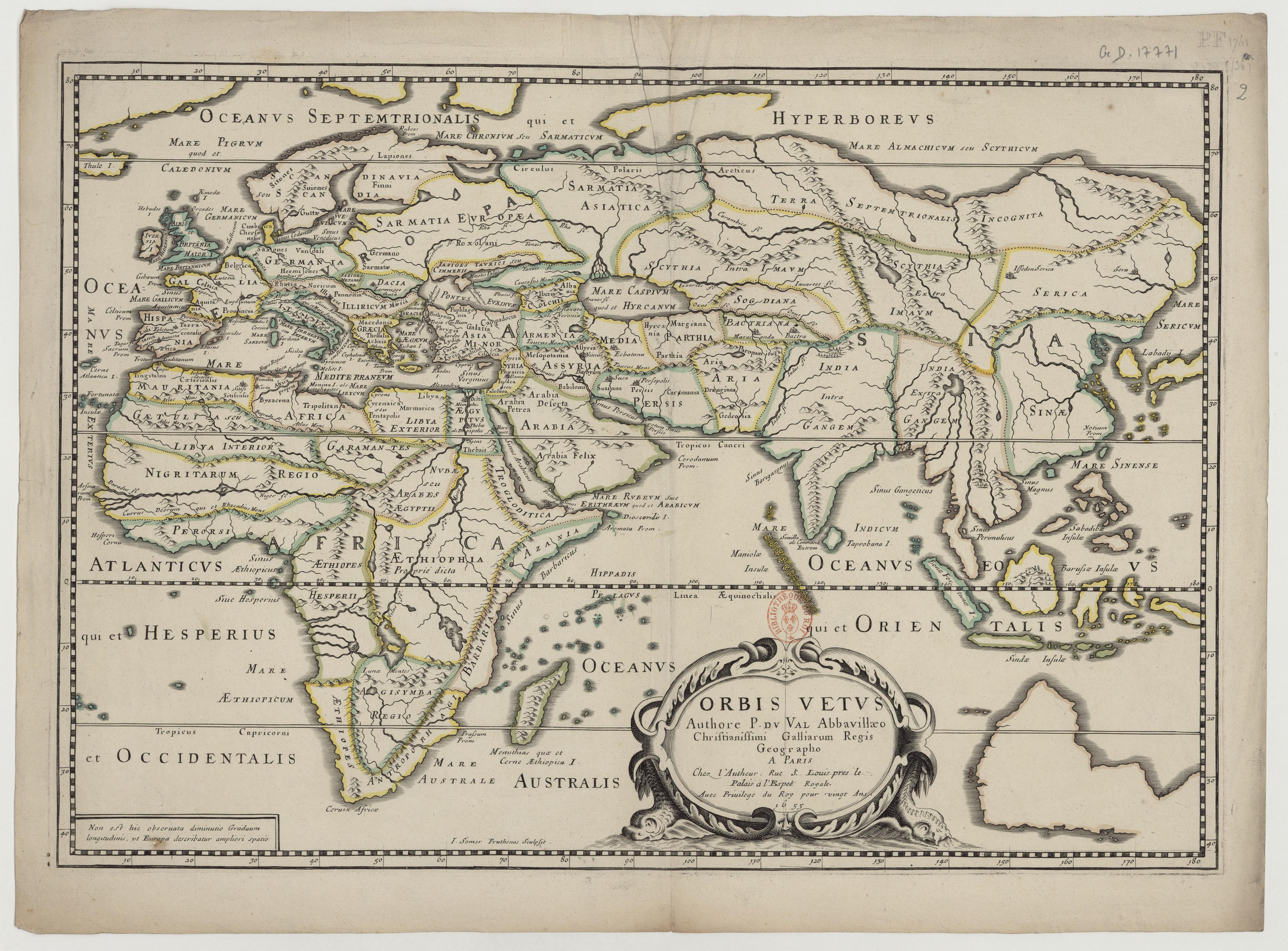}\\[.6em]
\includegraphics[width=\linewidth,height=.43\textheight,keepaspectratio]{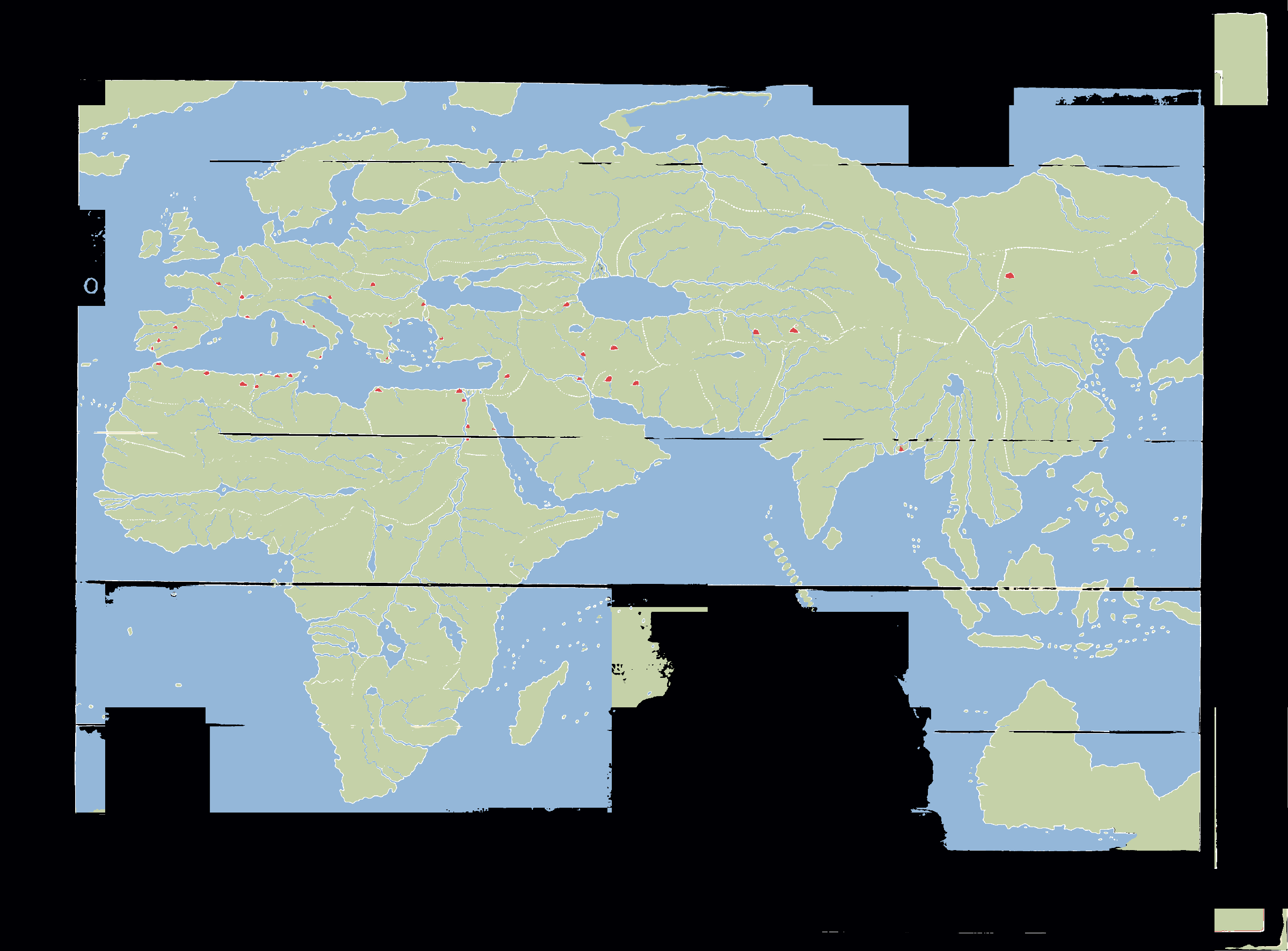}
\caption{(top) Map of the Old World, 1655. Pierre Duval. Orbis Vetus, 1655. Published in Paris. Copperplate, hand colored. 38.5 x 54 cm. BnF, GE D-17771. URL: \url{gallica.bnf.fr/ark:/12148/btv1b84953694}. (bottom) Result of the semantic segmentation. Seas and continents are correctly segmented despite their unusual shape.}
\label{fig:a1}
\end{figure}

\begin{figure}[pos=htbp,align=\centering]
\includegraphics[width=\linewidth,height=.43\textheight,keepaspectratio]{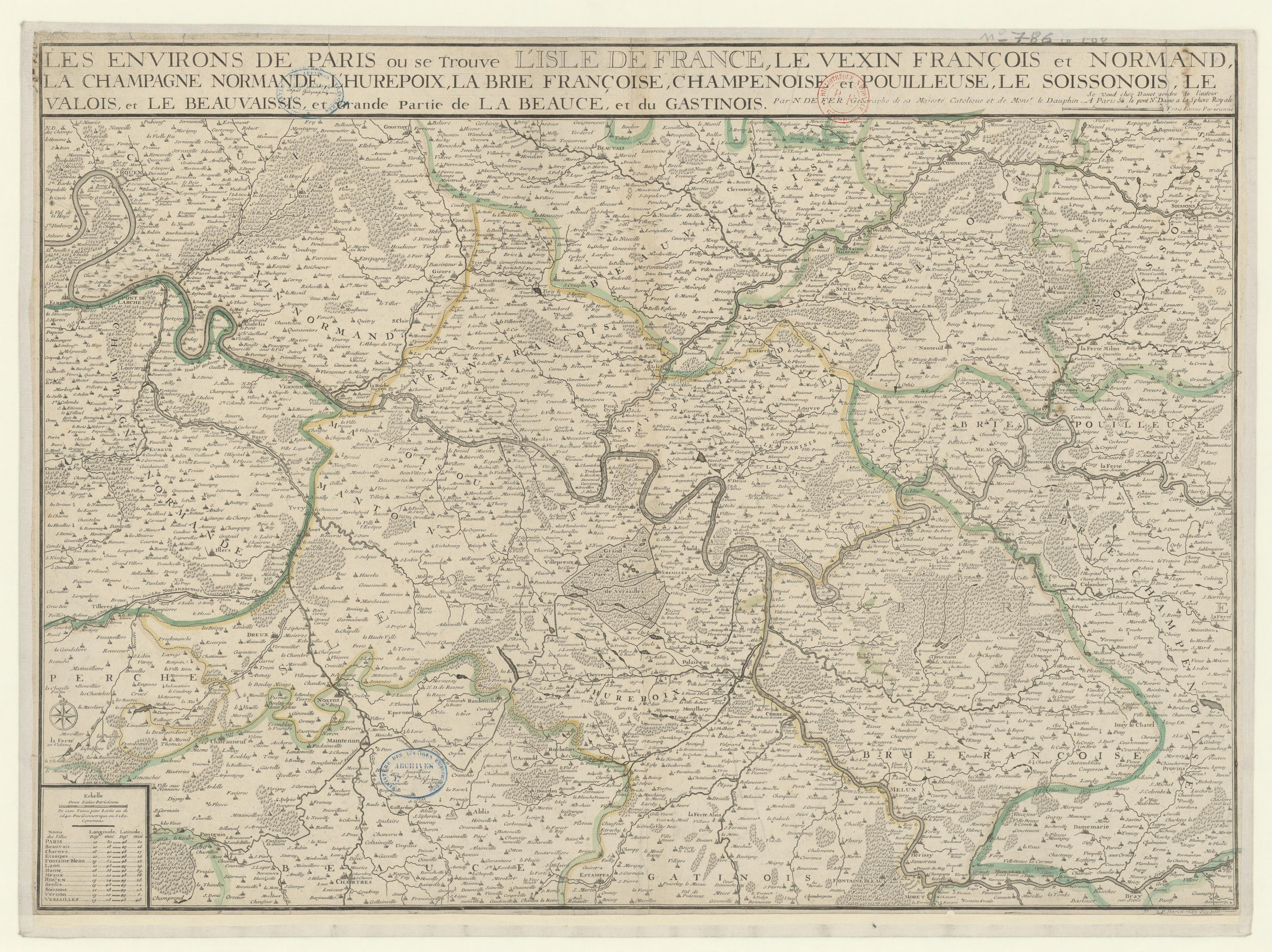}\\[.6em]
\includegraphics[width=\linewidth,height=.43\textheight,keepaspectratio]{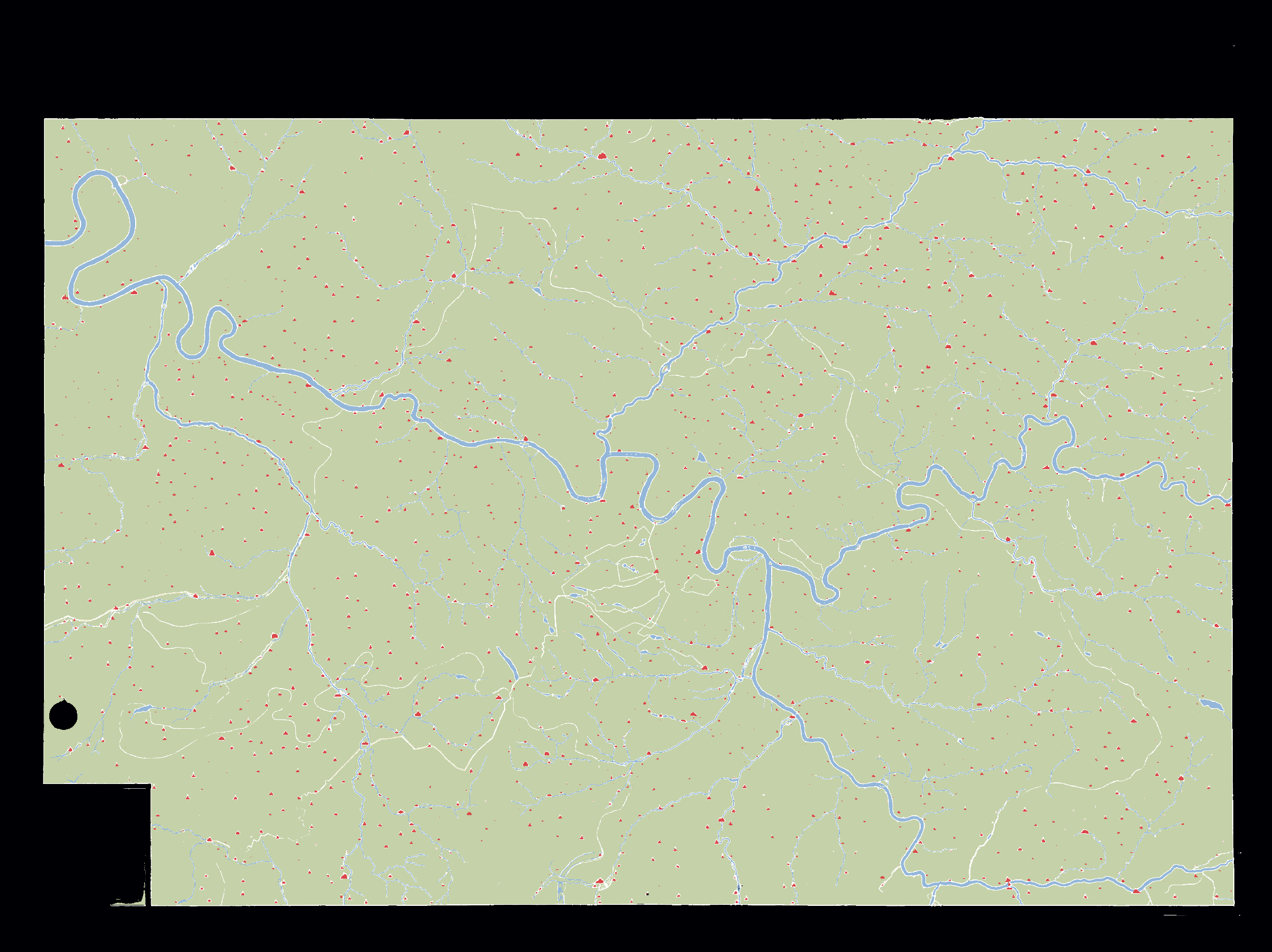}
\caption{(top) Map of Île-de-France region, 18th century. Nicolas de Fer, P. Starckman. Les environs de Paris [...], 18th century. Danet, Paris. Copperplate, hand colored. 48.5 x 67.5 cm. BnF, GE DD-2987 (786B). URL: \url{gallica.bnf.fr/ark:/12148/btv1b530532084}. (bottom) Result of the semantic segmentation. Waterways are well segmented; city icons are correctly classified as built.}
\label{fig:a2}
\end{figure}

\begin{figure}[pos=htbp,align=\centering]
\includegraphics[width=\linewidth,height=.43\textheight,keepaspectratio]{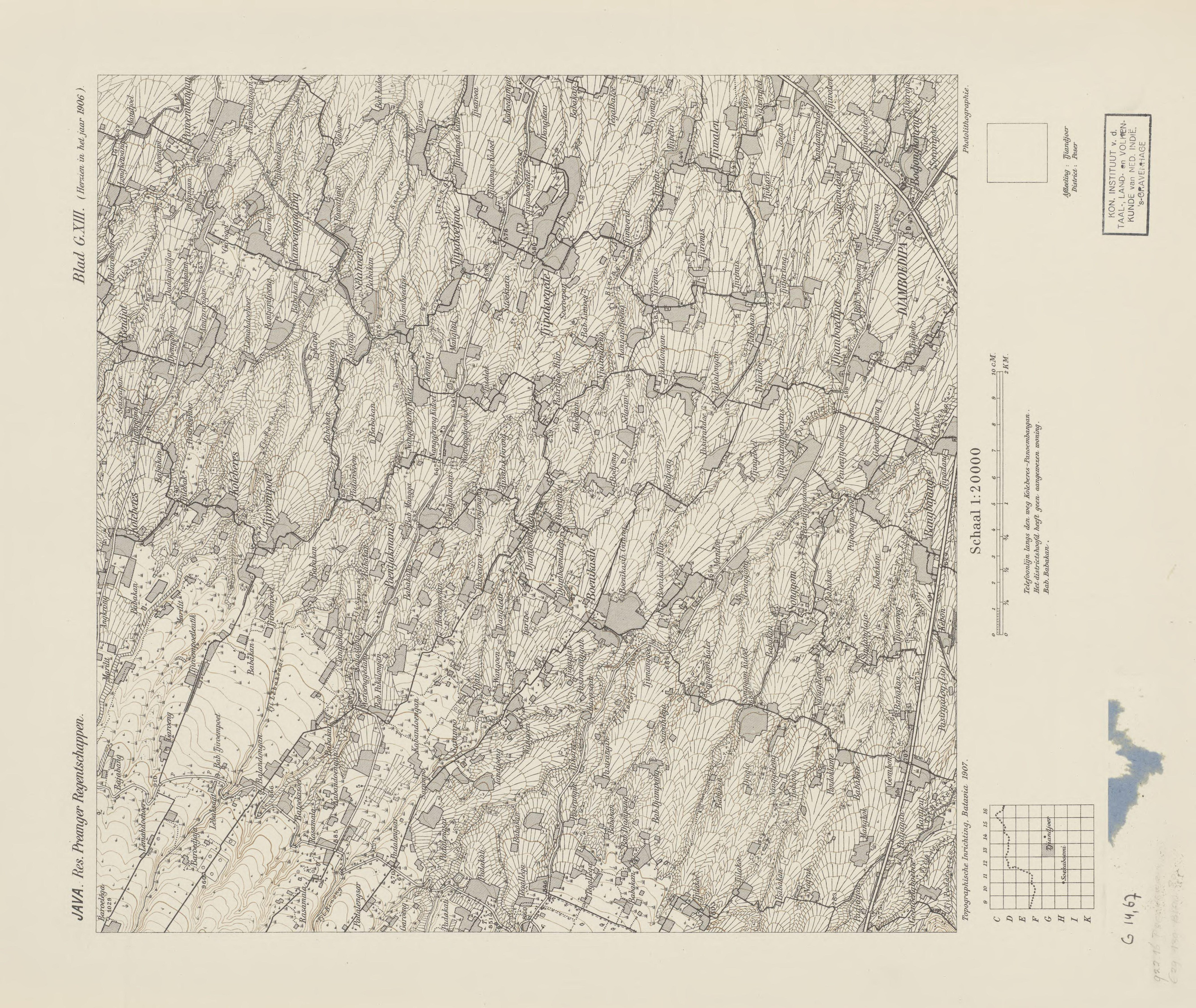}\\[.6em]
\includegraphics[width=\linewidth,height=.43\textheight,keepaspectratio]{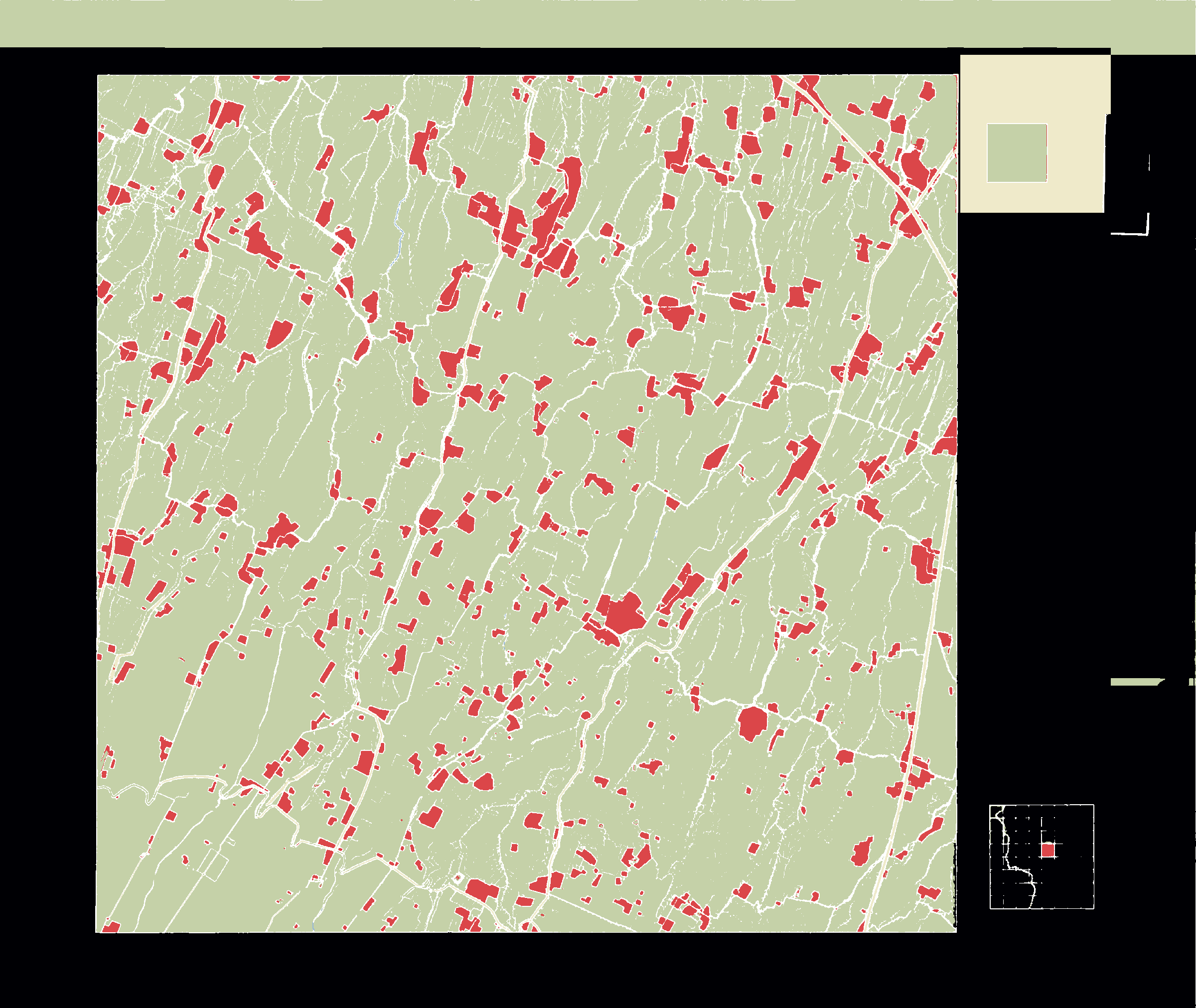}
\caption{(top) Jambudipa in the topographic map of Java at the scale of 1:20,000, 1907. Java Res Preanger Regentschappen, Blad G.XIII, 1907. Dutch Topographic Bureau, Batavia. Photoengraving. Leiden University Library, DG 14,67. Rotated 90°. URL: \url{hdl.handle.net/1887.1/item:815459}. (bottom) Result of the semantic segmentation. Urban areas are well segmented in spite of their unusual morphology; the recognition of boundaries seems impaired by the dense representation of relief.}
\label{fig:a3}
\end{figure}

\begin{figure}[pos=htbp,align=\centering]
\includegraphics[width=\linewidth,height=.43\textheight,keepaspectratio]{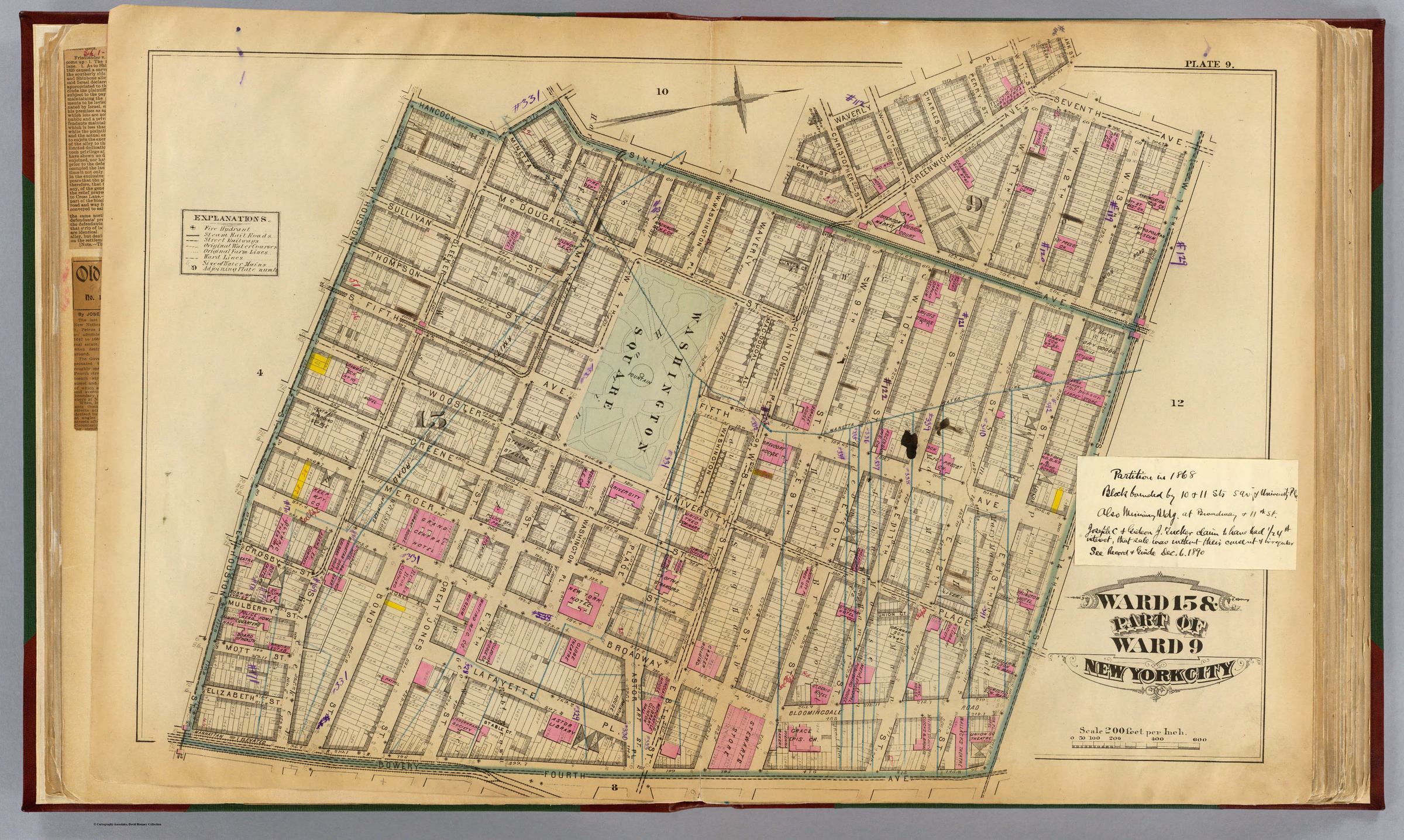}\\[.6em]
\includegraphics[width=\linewidth,height=.43\textheight,keepaspectratio]{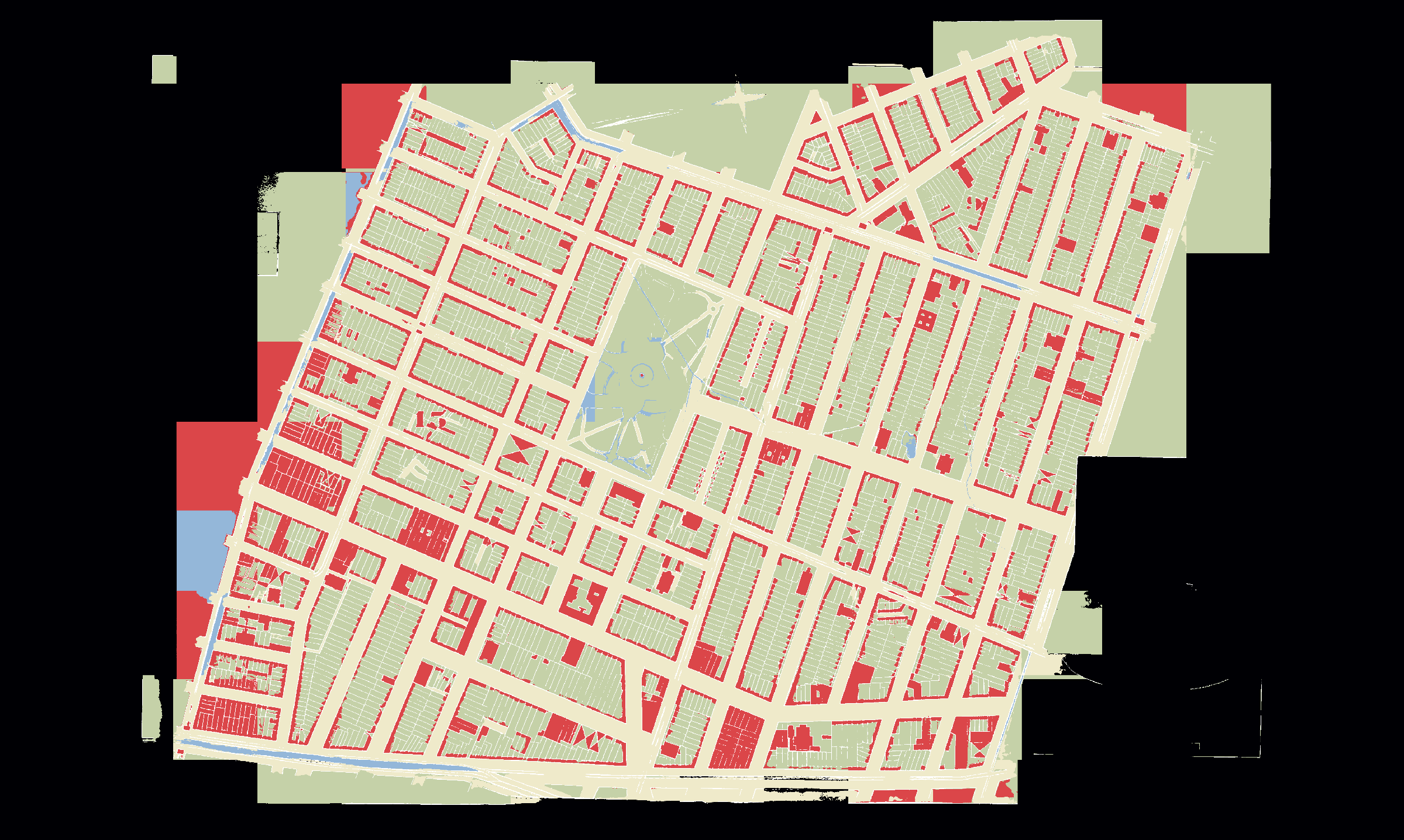}
\caption{(top) New York City Atlas, 1879. George W. Bromley, Edward Robinson, August H. Mueller. New York City, Ward 15 \& Part of Ward 9, 1879. Printed by F. Bourquin, Philadelphia. Published by G.W. Bromley \& Co, and E. Robinson, New York. Lithography, hand colored. 46 x 67 cm. David Rumsey Collection, 2597.010. URL: \url{davidrumsey.com/luna/servlet/detail/RUMSEY~8~1~30628~1150150}. (bottom) Result of the semantic segmentation. The street grid and land plots tend to be well segmented, although disruptions in the urban form and the limited context in directly adjacent blank areas generate confusion.}
\label{fig:a4}
\end{figure}

\begin{table}[pos=htbp]
\centering
\scriptsize
\begin{adjustbox}{max width=\linewidth}
\begin{tabular}{@{}p{.07\linewidth}p{.21\linewidth}p{.13\linewidth}rrp{.35\linewidth}@{}}
\toprule
Class* & Feature & Category & Zoom min & Zoom max & Query filter \\
\midrule
built & landuse\_\allowbreak{}residential & landuse & 10 & 11 & (class IN ('residential','suburb','neighborhood')) \\
nblt1 & landcover\_\allowbreak{}grass & landcover &  &  & (class IS 'grass') \\
nblt2 & landcover\_\allowbreak{}wood & landcover &  &  & (class IS 'wood') \\
nblt1 & landcover\_\allowbreak{}sand & landcover &  &  & (class IS 'sand') \\
nblt1 & landcover\_\allowbreak{}glacier & landcover &  &  & (subclass IN ('glacier','ice\_\allowbreak{}shelf')) \\
water & water & water &  &  & NOT (intermittent) AND (brunnel IS NOT 'tunnel') \\
water & water\_\allowbreak{}intermittent & water &  &  & (intermittent) \\
water & waterway & waterway &  &  & (brunnel IS null) OR (brunnel NOT IN ('tunnel','bridge') AND NOT (intermittent)) \\
built & building & building & 13 &  & (all) \\
roadn1 & road\_\allowbreak{}area\_\allowbreak{}pier & transportation &  &  & (\_\allowbreak{}geom\_\allowbreak{}type IS Polygon) AND (class IS 'pier') \\
roadn1 & road\_\allowbreak{}area\_\allowbreak{}bridge & transportation &  &  & (\_\allowbreak{}geom\_\allowbreak{}type IS Polygon) AND (brunnel IS 'bridge') \\
roadn1 & road\_\allowbreak{}pier & transportation & 14 &  & (class IN ('pier')) \\
roadn1 & road\_\allowbreak{}minor & transportation & 13 &  & (class IN ('minor','service')) \\
roadn1 & road\_\allowbreak{}major & transportation &  &  & (class IS 'motorway') \\
roadn1 & road\_\allowbreak{}motorway & transportation & 4 &  & (class IS 'motorway') \\
roadn2 & railway & transportation & 11 &  & (class IS 'rail') \\
roadn1 & bridge & transportation &  &  & (brunnel IS 'bridge') AND (class IN ('primary','secondary','tertiary')) \\
bnd & admin\_\allowbreak{}sub & boundary & 3 &  & ('admin\_\allowbreak{}level' IN (4,6,8)) \\
built & (all) & place & 3 & 9 & (class IN ('city','town')) \\
text & label\_\allowbreak{}airport & aerodrome\_\allowbreak{}label & 10 &  & ('iata' IS NOT null) \\
text & label\_\allowbreak{}road & transportation\_\allowbreak{}name & 13 &  & (\_\allowbreak{}geom\_\allowbreak{}type IS LineString) AND (subclass IS NOT 'ferry') \\
text & label\_\allowbreak{}place\_\allowbreak{}other & place & 8 &  & (\_\allowbreak{}geom\_\allowbreak{}type IS Point) AND (class IS null OR class NOT IN ('city','state','country','continent')) \\
text & label\_\allowbreak{}place\_\allowbreak{}city & place &  & 16 & (\_\allowbreak{}geom\_\allowbreak{}type IS Point) AND (class IS 'city') \\
text & label\_\allowbreak{}country\_\allowbreak{}other & place &  & 12 & (\_\allowbreak{}geom\_\allowbreak{}type IS Point) AND (class IS 'country') \\
text & label\_\allowbreak{}water & water\_\allowbreak{}name & 10 &  & (\_\allowbreak{}geom\_\allowbreak{}type IN (Polygon,LineString)) \\
\bottomrule
\end{tabular}

\end{adjustbox}
\caption{List of vector objects queried from MapTiler Planet API. The feature and category fields correspond to MapTiler's data structure. The filter field is adapted for legibility. }
\label{tab:a4}
\par\smallskip\raggedright * nblt = non-built (1 base, 2 forest), roadn = road network (1 base, 2 railway), bnd = boundary.
\end{table}

\clearpage

\end{document}